\newif\ifsingle

\newif\ifproofs

\ifsingle
\documentclass[11pt,draftclsnofoot, onecolumn]{IEEEtran}		
\else		
\documentclass[10pt,journal,comsoc]{IEEEtran} 
\fi

\usepackage{times}
\usepackage{amsmath,dsfont}
\usepackage{bbm}
\usepackage{amssymb,amsthm}
\usepackage{epsfig,verbatim} 
\usepackage{setspace}
\usepackage{color}
\usepackage{xcolor}
\usepackage{cite}
\usepackage{epstopdf}
\usepackage{graphics}
\usepackage{graphicx}
\usepackage{accents}
\usepackage{acronym}
\usepackage[bookmarks,colorlinks]{hyperref}
\usepackage{booktabs}
\usepackage{mathtools}
\usepackage{dblfloatfix}
\usepackage{float}
\usepackage[linesnumbered,ruled,vlined]{algorithm2e}
\usepackage{algpseudocode}
\usepackage{enumitem}
\usepackage{bm}
\usepackage{placeins}
\usepackage{tablefootnote}
\usepackage{threeparttable}
\usepackage{subcaption}

\newcommand{\removelatexerror}{\let\@latex@error\@gobble}
\newcommand{\myVec}[1]{{\boldsymbol{#1}}}
\newcommand{\myMat}[1]{{\boldsymbol{#1}}}
\newcommand{\mySet}[1]{\mathcal{#1}}

\let\oldnl\nl
\newcommand{\nonl}{\renewcommand{\nl}{\let\nl\oldnl}}

\acrodef{fc}[FC]{fully-connected}
\acrodef{bc}[BC]{broadcast channel}
\acrodef{mac}[MAC]{multiple access channel}  
\acrodef{adc}[ADC]{analog-to-digital converter}  
\acrodef{csi}[CSI]{channel state information} 
\acrodef{snr}[SNR]{signal-to-noise ratio}
\acrodef{sinr}[SINR]{signal-to-interference-and-noise ratio}
\acrodef{tx}[TX]{Transmitter}  
\acrodef{mimo}[MIMO]{multiple-input multiple-output}
\acrodef{mse}[MSE]{mean-squared error}
\acrodef{lmmse}[LMMSE]{linear minimum mean-squared error}
\acrodef{pdf}[PDF]{probability density function}
\acrodef{rv}[RV]{random variable} 
\acrodef{isi}[ISI]{intersymbol interference}  
\acrodef{awgn}[AWGN]{additive white Gaussian noise} 
\acrodef{lti}[LTI]{linear time-invariant}  
\acrodef{ut}[UT]{user terminal} 
\acrodef{mmw}[mmWave]{millimeter wave}
\acrodef{dma}[DMA]{dynamic metasurface antenna}
\acrodef{ofdm}[OFDM]{orthogonal frequency division multiplexing}
\acrodef{ofdma}[OFDMA]{orthogonal frequency division multiple access}
\acrodef{dnn}[DNN]{deep neural network}
\acrodef{gnn}[GNN]{graph neural network}
\acrodef{ml}[ML]{machine learning}
\acrodef{dl}[DL]{deep learning}
\acrodef{mlp}[MLP]{multilayer perceptron}
\acrodef{sgd}[SGD]{stochastic gradient descent}
\acrodef{bpsk}[BPSK]{binary phase shift keying}
\acrodef{pgd}[PGD]{projected gradient descent}
\acrodef{ber}[BER]{bit error rate}
\acrodef{csi}[CSI]{channel state information}
\acrodef{map}[MAP]{maximum a-posteriori probability}
\acrodef{ct}[CT]{continuous-time}
\acrodef{manet}[MANET]{mobile ad hoc network}
\acrodef{noma}[NOMA]{non-orthogonal multiple access}
\acrodef{sic}[SIC]{successive interference cancellation}
\acrodef{mmse}[MMSE]{minimum mean squared error}
\acrodef{gat}[GAT] {Graph Attention Network}
\acrodef{adam}[ADAM]{Adaptive Moment Estimation}
\acrodef{film}[FiLM]{Feature-wise Linear Modulation}
\acrodef{jk}[JK]{Jumping Knowledge}
\acrodef{siso}[SISO]{Single-Input-Single-Output}
\acrodef{dwp}[DWP]{Distributed Widest Path}
\acrodef{dgr}[DGR]{Distributed Greedy Split}
\acrodef{ffn}[FFN]{Feedforward Network}
\acrodef{cwp}[CWP]{Centralized Widest Path}
\acrodef{cgs}[CGS]{Centralized Greedy Split}

\definecolor{blue}{rgb}{0,0,1}

\DeclareMathOperator*{\argmax}{argmax}

\ifsingle

\else

\fi 

\usepackage[all=normal,paragraphs=tight,floats=normal,mathspacing=normal,wordspacing=tight,charwidths=tight,mathdisplays=normal,leading=normal]{savetrees}

\IEEEoverridecommandlockouts

\title{MANET-GNN: Learned Decentralized Optimization of Power Allocation in Multi-Channel MANETs 
}
\author{
	\IEEEauthorblockN{Tomer Alter, Nir Shlezinger, and Michael Segal\\
    \thanks{ Parts of this work were presented at the IEEE International Conference on Communications (ICC) 2026 as the paper \cite{alter2026decentralized}.
The authors are with the ECE School, Ben-Gurion University of the Negev, Israel (e-mail: tomeralt@post.bgu.ac.il; \{nirshl; segal\}@bgu.ac.il).  The work was supported  by the European Research Council (ERC) under the ERC starting grant nr. 101163973 (FLAIR), by  Israeli Science Foundation (Grant No. 465/22) and
US Army Research Office under Grant Number W911NF-22-1-0225.
}
	}

	\vspace{-0.5cm}
	
}
\allowdisplaybreaks
\begin{document}

\maketitle 
\begin{abstract}  
\Acp{manet} enable flexible infrastructure-less wireless connectivity in dynamic and resource-constrained environments. As modern \acp{manet} exploit multiple frequency channels and support heterogeneous traffic patterns, decentralized transmit-power allocation becomes increasingly challenging. 
We develop a unified learned optimization framework for decentralized power allocation in dynamic multi-hop, multi-channel \acp{manet}. We formulate a constrained end-to-end throughput maximization problem covering unicast, multicast, multicommodity, convergecast, and many-to-many communication. Although centralized and non-convex, this problem serves as an unsupervised training objective for \emph{MANET-GNN}, a message-passing \ac{gnn} that operates as a distributed learned optimizer. MANET-GNN uses only local, possibly noisy, \ac{csi} and a prescribed number of neighbor message exchanges, enabling low-latency decentralized inference while generalizing across topologies and network sizes. Numerical results show that MANET-GNN achieves centralized-competitive performance across communication frameworks, remains robust to channel uncertainty, and scales effectively across \ac{manet} configurations.
\end{abstract}

\acresetall

\section{Introduction}

\Acp{manet} are infrastructure-less wireless networks in which mobile devices autonomously form multi-hop topologies and sustain end-to-end connectivity~\cite{tavli2006mobile}. 
They arise in applications such as vehicular coordination, industrial IoT, public safety, and temporary communication infrastructures, where nodes are energy-limited, latency-constrained, and subject to rapidly varying topologies and channels. 
Efficient and scalable resource allocation is essential for exploiting the potential of \acp{manet}~\cite{kafetzis2022software}.
Modern \ac{manet} technologies increasingly support \emph{multi-channel} communication, where each link may use several orthogonal channels, e.g., through heterogeneous technologies, multi-band radios, or multi-carrier signaling~\cite{xie2021multi,karabulut2022novel,chen2024joint}. 
While this provides additional degrees of freedom for improving throughput and reliability, it also complicates resource allocation: nodes must distribute limited transmit power across outgoing links, channels, and possibly multiple concurrent messages, while accounting for multi-hop routing and end-to-end performance. 
This motivates scalable decentralized mechanisms for joint routing-aware power allocation in multi-channel \acp{manet}.

\begin{figure}
   \centering
  \includegraphics[width=0.85\columnwidth,keepaspectratio]{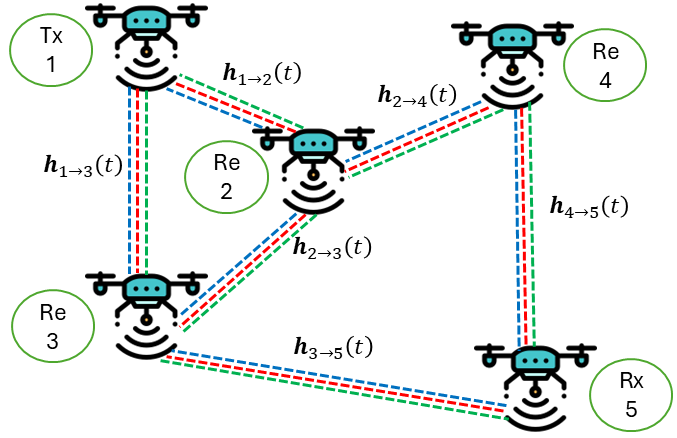}
  \caption{Illustration of a multi-channel \ac{manet} with $B=3$ channels, $|\mySet{V}|=5$ nodes, and $|\mySet{E}|=6$ links.}
  \label{fig:example graph}
\end{figure}

A broad range of optimization-based methods has been proposed for routing and power allocation in \acp{manet}~\cite{kanellopoulos2020survey}, including shortest- and widest-path routing, backpressure scheduling, max--min fair allocation~\cite{alotaibi2012survey,jayakumar2007ad,mishra2013power}, radio-state adaptation~\cite{singh1998pamas,ye2002smac}, load balancing~\cite{kim2002power}, location- and multicast-based routing~\cite{karp2000gpsr,li2001learn}, and cross-layer optimization~\cite{mafirabadza2016efficient}. 
Learning-based routing and resource-allocation methods have also been considered for heterogeneous wireless networks with multiple flows~\cite{kim2025deep, kim2025deepb, kong2025joint, haque2025deer, gillani2025optimal, kong2024decentralized}. 
However, existing approaches typically focus on single-channel systems, specific traffic patterns, or routing without joint decentralized power allocation, and therefore do not directly address multi-channel \acp{manet} with diverse communication objectives.

Data-driven methods offer an alternative to repeatedly solving difficult optimization problems. Distributed reinforcement-learning approaches equip agents with local \acp{dnn} for decentralized resource allocation~\cite{liu2024graph,dai2024survey,feng2025gnn}, but may not explicitly exploit graph structure, limiting their generalization across topologies. 
The \emph{learn-to-optimize} paradigm trains neural networks or unfolded optimizers to rapidly generate feasible solutions~\cite{shlezinger2022model,shlezinger2025deep}, with applications to centralized resource allocation~\cite{alter2025rapid} and decentralized consensus optimization~\cite{noah2024distributed,saravanos2024deep}.

For graph-structured decentralized problems, \acp{gnn} provide a natural architecture due to their permutation equivariance~\cite{bronstein2021geometric} and local message-passing operations~\cite{gilmer2017neural,kipf2016semi,feng2022powerful}. 
They have been applied to wireless power allocation~\cite{eisen2020optimal,randall2022grows,yang2024knowledge}, resource allocation~\cite{peng2024learning}, graph-unfolded optimization~\cite{yang2024knowledge}, over-the-air graph learning~\cite{gu2023graph}, link scheduling~\cite{zhao2022link,zhao2025distributed}, routing~\cite{das2025opportunistic}, beamforming~\cite{shen2020graph}, and wireless federated learning~\cite{gao2025graph}. 
Nevertheless, existing works generally target individual tasks in infrastructure-based or interference-limited networks rather than decentralized multi-hop, multi-channel \acp{manet} supporting multiple communication frameworks. This gap motivates the proposed \ac{manet}-\ac{gnn} architecture.

Building on these insights, we develop a unified learned optimization framework for decentralized power allocation in multi-channel \acp{manet}. 
We first formulate routing-aware power allocation as a constrained optimization problem that maximizes an end-to-end throughput metric encompassing several \ac{manet} communication frameworks~\cite{el2011network}. 
Although the resulting problem is centralized and non-convex, it serves as an unsupervised training objective for a dedicated \ac{gnn} architecture, termed \emph{MANET-GNN}. 
The learned policy operates using only local \ac{csi} and a limited number of message exchanges, while approximating centralized optimization across varying topologies and communication objectives.

Our main contributions are:
\begin{itemize}
    \item \textbf{Unified formulation for multi-framework \acp{manet}:} 
    We formulate multi-channel \ac{manet} power allocation for unicast, multicast, multicommodity, convergecast, and many-to-many scenarios~\cite{el2011network} using a common end-to-end rate objective.

    \item \textbf{MANET-GNN decentralized learned optimizer:} 
    We design a message-passing \ac{gnn} that respects decentralization and latency constraints, uses only local possibly noisy \ac{csi}, and generalizes across topologies and network sizes.

    \item \textbf{Framework-aware unsupervised training:} 
    We train MANET-GNN directly through the centralized optimization objective, without requiring ground-truth power allocations, while accounting for the communication framework and \ac{csi} uncertainty.

    \item \textbf{Extensive experimentation:} 
    We show that MANET-GNN achieves near-centralized performance across multiple communication frameworks, remains robust to \ac{csi} uncertainty, and scales to unseen network sizes and topologies.
\end{itemize}

The remainder of the paper is organized as follows. Section~\ref{sec:System} presents the system model and problem formulation. Section~\ref{sec:Decentralized} introduces MANET-GNN, and Section~\ref{sec: Experimental} evaluates its performance. Section~\ref{sec:Conclusions} concludes the paper.

Throughout the paper, $\|\cdot\|_0$ denotes the $\ell_0$ pseudo-norm, i.e., the number of nonzero entries. 
The Frobenius norm is denoted by $\|\cdot\|_F$. 
For tensor indexing, $[\myMat{X}]_{i_1,i_2,\dots,i_n}$ denotes a scalar entry, while $[\myMat{X}]_{:,\dots,:,i_k,:,\dots,:}$ denotes the sub-tensor obtained by fixing the $k$th index.
 
\vspace{-0.1cm}
\section{System Model}
\label{sec:System}
In this section, we formulate the system model for multi-channel \acp{manet}. We commence with presenting the \ac{manet} system in Subsection~\ref{sec:Com system}, and  the different communication frameworks considered for such multi-user networks in Subsection~\ref{sec:Comm frameworks}.  
Based on these, we 
formulate the decentralized power allocation optimization problem in Subsection~\ref{sec:Problem}. 

\vspace{-0.1cm}
\subsection{MANET System Model}
\label{sec:Com system}

We consider a dynamic multi-hop multi-channel \ac{manet} with reciprocal links, as illustrated in Fig.~\ref{fig:example graph}. 
The dynamic nature implies that both the \ac{manet} topology and the corresponding channels can change in a block-wise fashion. Specifically, during block $t$, the network topology is modeled as an undirected connected graph $\mathcal{G}(t) = (\mathcal{V}(t),\mathcal{E}(t))$, where $\mathcal{V}(t)$  is the set of nodes (user devices) and $\mathcal{E}(t) \subseteq \mathcal{V}(t)\times\mathcal{V}(t)$ is the set of  links. The dependence on $t$ indicates that users can join and leave the network, and that  connectivity can change due to, e.g., user mobility. 


Each link in the multi-channel \ac{manet} can utilize one of $B$ communication resources, e.g., frequency channels, subcarriers, or resource blocks. The considered variations are described as block-fading, where the coefficients of the channels are assumed to remain constant within a block and vary independently across blocks, capturing the temporal dynamics of mobile ad hoc environments~\cite{biglieri2001limiting, yang2013block}. Specifically, during block $t$, the channel over link $(i,j)$ is represented by a $B\times 1$ vector
$
\mathbf{h}_{i\leftrightarrow j}(t)
\triangleq
\left[
h_{i\to j}^{(1)}(t),\ldots,h_{i\to j}^{(B)}(t)
\right]^{\top}\in\mathbb{C}^B,
$
where $h_{i\to j}^{(b)}(t)$ denotes the channel realization over channel $b$. Reciprocity implies that for every
$(i,j)\in\mathcal{E}(t)$, the forward and reverse channels are
identical.

The considered formulation also accommodates heterogeneous
communication capabilities across nodes. Specifically, a node is
not required to support all $B$ communication resources. If node
$i$ does not support resource $b$, the corresponding channel
coefficients are set to zero, i.e., $h_{i\to j}^{(b)} = 0$ for all
$j\in\mathcal{V}$. Consequently, the corresponding links are
removed from the feasible topology on resource $b$, allowing the
model to represent heterogeneous communication technologies and
frequency availability across the network.

Let $\mySet{N}_j(t)$ denote the set of one-hop neighbors of
node $j$ in $\mathcal{G}(t)$, i.e.,
$\mySet{N}_j(t)=\{l\in\mathcal{V}(t):(l,j)\in\mathcal{E}(t)\}$.
Transmissions over different channels are orthogonal, while
simultaneous transmissions over the same channel may interfere
at neighboring receivers. Thus, letting $s_{i\to j}^{(b)}(t)\in
\mathbb{C}$ be the unit-variance signal transmitted by node $i$
to node $j$ over channel $b$, node $j$ observes
\begin{equation}
\label{eq:received message b}
y_{i\to j}^{(b)}(t)
=
h_{i\to j}^{(b)}(t)
p_{i\to j}^{(b)}(t)
s_{i\to j}^{(b)}(t)
+
I_j^{(b)}(t)
+
w_j^{(b)}(t),
\end{equation}
where  $I_j^{(b)}(t)$, defined as
\begin{align*}
I_j^{(b)}(t)
=
\sum_{l\in\mySet{N}_j(t)\setminus\{i\}}
h_{l\to j}^{(b)}(t)
p_{l\to j}^{(b)}(t)
s_{l\to j}^{(b)}(t),
\end{align*}
 represents the aggregate co-channel interference
at node $j$ generated by neighboring transmitters that reuse
channel $b$, and
$w_{j}^{(b)}(t)\sim \mySet{CN}(0,\sigma_b^2)$ is \ac{awgn}~\cite{tse2005fundamentals}.
For convenience, the key variables used in the system model are summarized in Table~\ref{tab:variables}.  

\begin{figure*}[t]
    \centering
    \includegraphics[width=0.8\textwidth]{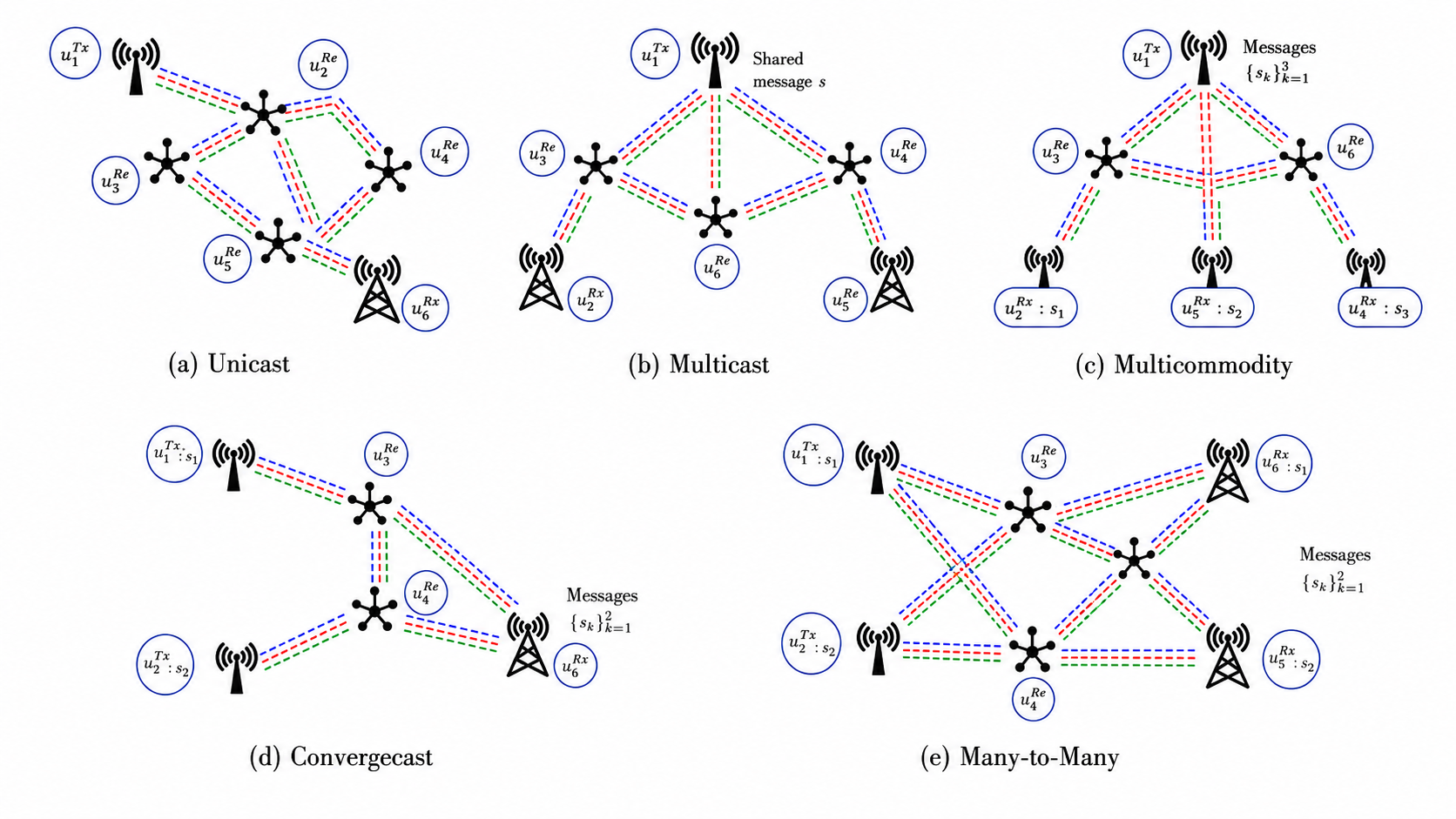}
    \caption{Illustration of the considered multi-channel \ac{manet}
    communication frameworks.}
    \label{fig:Frameworks}
\end{figure*}

\vspace{-0.1cm}
\subsection{Communication Frameworks over MANETs}
\label{sec:Comm frameworks}
The dynamic multi-channel \ac{manet} detailed in Subsection~\ref{sec:Com system} represents a multi-hop multi-user communication network. As such, it accommodates various communication frameworks which, as illustrated in Fig.~\ref{fig:Frameworks},
vary based on the specification of the set of source nodes, destination nodes, and message relations~\cite{el2011network}. We focus here on five different representative communication frameworks that can be supported by a multi-channel \acp{manet} with topology $\mathcal{G}=(\mySet{V},\mySet{E})$:
\begin{enumerate}[label={F\arabic*}]
\item \label{item:unicast} {\em Unicast}, which involves a single transmitter--receiver pair. A designated node $u^{\rm Tx}\in \mySet{V}$ acts as the source and a node $u^{\rm Rx} \in \mySet{V}$ acts as the destination, while all other nodes operate as relays. The objective is to deliver a single message from $u^{\rm Tx}$ to $u^{\rm Rx}$ via multi-hop routing.  
\item \label{item:multicast} {\em Multicast}, where a single transmitter $u^{\rm Tx}\in \mySet{V}$ communicates a common message to a set of $Q$ receivers $\{u^{\rm Rx}_1,\ldots,u^{\rm Rx}_Q\} \subset \mySet{V}$. All receivers expect to decode the same information. This framework generalizes broadcast-style dissemination with the requirement that each destination successfully receives the common message.
\item \label{item:multicom} {\em Multicommodity}, which extends multicast by associating each receiver with its own message. A single transmitter $u^{\rm Tx}\in \mySet{V}$ sends $K$ independent messages to $K$ receivers $\{u^{\rm Rx}_1,\ldots,u^{\rm Rx}_K\}\subset \mySet{V}$, where receiver $u^{\rm Rx}_k$ expects the $k$th message $m_k$. Unlike multicast, messages are distinct and cannot be reused across destinations. This setting introduces traffic differentiation and competition for network resources.
\item \label{item:convergecast} {\em Convergecast}, which can be viewed as the inverse of the multicommodity framework. Here, a set of transmitters $\{u^{\rm Tx}_1,\ldots,u^{\rm Tx}_K\}\subset \mySet{V}$ each hold a distinct message, which must be delivered to a single receiver $u^{\rm Rx}\in \mySet{V}$. This setting arises naturally in data gathering, sensor fusion, and consensus applications, where information from multiple sources must be collected at a single node.

\item \label{item:many2many} {\em Many-to-Many}, which extends multicommodity by allowing multiple transmitters to simultaneously inject independent traffic into the network. Specifically, a set of transmitters
$\{u^{\rm Tx}_1,\ldots,u^{\rm Tx}_K\}\subset\mathcal{V}$
communicates with a set of receivers
$\{u^{\rm Rx}_1,\ldots,u^{\rm Rx}_K\}\subset\mathcal{V}$,
where transmitter $u^{\rm Tx}_k$ sends message $m_k$ to receiver
$u^{\rm Rx}_k$. Unlike multicommodity communication, messages may originate from different source nodes, resulting in multiple concurrent communication flows that compete for network resources and jointly determine the routing and power-allocation decisions.
\end{enumerate}

While \ref{item:unicast}-\ref{item:many2many} differ in message relationships and traffic patterns, they are all evaluated over the same physical-layer model described in Subsection~\ref{sec:Com system}. The distinction arises at the network layer: information must be routed from transmitters to receivers over multi-hop paths in $\mathcal{G}$. In this work, routing is handled at the optimization layer by explicitly evaluating available paths between communicating nodes and selecting paths based on their achievable end-to-end performance, as formalized next.

\begin{table}
    \centering
    \caption{Key variables and parameters}
    \label{tab:variables}
    \resizebox{\columnwidth}{!}{%
    \begin{tabular}{ll}
        \hline
        \textbf{Symbol} & \textbf{Definition} \\
        \hline
        $w_{{j}}^{(b)}(t)$ & AWGN noise received at node $j$ on channel $b$\\
        $h_{i \rightarrow j}^{(b)}(t)$ & Channel coefficient between nodes $i$ and $j$ on channel $b$ \\
        $p_{i \rightarrow j}^{(b)}(t)$ & Power allocated by node $i$ to node $j$ on channel $b$ \\
        $s_i^{(b)}(t)$ & Transmitted signal from node $i$ on channel $b$ \\
        $\mySet{N}_j(t)$ & Set of neighboring nodes of node $j$ \\
        \hline
    \end{tabular}%
    }
\end{table}

\vspace{-0.1cm}
\subsection{Power Allocation Problem Formulation}
\label{sec:Problem}
The ability to reliably carry out the communication frameworks \ref{item:unicast}-\ref{item:many2many} largely depends on how the nodes allocate their transmission power across the $B$ channels for all links 
in the \ac{manet}. Namely, power allocation refers to  setting  the collection $\{ p_{i \rightarrow j}^{(b)}(t)\}$ for all $(i,j)\in \mySet{E}(t)$ and $b\in\{1,\ldots,B\}$, such that appropriate end-to-end performance metrics are optimized under \ref{item:unicast}-\ref{item:many2many}, while satisfying per-node power constraints.

We assume that a message cannot switch channels between hops within a transmission block. Thus, all links carrying a message on channel $b$ use the same channel throughout its route. Per-hop channel switching would require additional coordination and channel-assignment variables, substantially increasing the optimization complexity and signaling overhead~\cite{mohsenian2007joint}. We leave dynamic channel adaptation for future work.

We next formulate the constraints on power allocation and cast this task as an optimization setting, based on which we introduce a unified formulation the power allocation problem. 
Since our formulation focuses a single signal block, we henceforth omit the block index $t$ for brevity.

\subsubsection{Power Constraint}
The communication frameworks \ref{item:unicast}-\ref{item:many2many} require each user to allocate its power between $K$ messages (with $K=1$ for single-message frameworks \ref{item:unicast}-\ref{item:multicast}) and across all its outgoing channels.  Accordingly, we stack the power allocation as a $B\times K \times |\mySet{V}| \times |\mySet{V}|$ tensor $\myMat{P}$, with
\begin{equation}
    \sum_{k=1}^K[\myMat{P}]_{b,k,i,j} =  p_{i \rightarrow j}^{(b)}.
\end{equation}

The variables $P_{b,k,i,j}$ denote transmission amplitudes, such that the corresponding received signal power is proportional to $P_{b,k,i,j}^{2}$. We assume orthogonal message transmission over each link--channel resource; hence,
\begin{equation}
    \|\mathbf{P}_{b,:,i,j}\|_0 \leq 1,
    \qquad
    (i,j)\in\mathcal{E},\quad b\in\{1,\ldots,B\}.
\end{equation}
Thus, at most one message occupies a given link and channel, although different links may reuse the same channel and consequently generate co-channel interference.

The transmit-power constraint is imposed jointly over all messages, channels, and outgoing links of each node. Since $\mathbf{P}$ contains amplitudes, it is expressed as
\begin{equation}
    \left\|[\mathbf{P}]_{:,:,i,:}\right\|_{F}
    =
    \sqrt{
    \sum_{b,k,j}
    \left(P_{b,k,i,j}\right)^2
    }
    \leq 1,
    \qquad \forall i\in\mathcal{V}.
\end{equation}
Moreover, each node is subject to a unit power constraint and cannot allocate negative power to transmission. As a result, the set of feasible power allocations can be written as
\begin{align}
\mySet{P}
= \big\{&
\myMat{P} \in [0,1]^{B \times K \times |\mySet{V}| \times |\mySet{V}|}
: \| [\myMat{P}]_{:,:,i,:} \|_{F} \le 1,\ \forall i \in \mySet{V}, 
\notag \\
&\| [\myMat{P}]_{b,:,i,j} \|_{0} \le 1,\ \forall(i,j,b)\in \mySet{E}\times\{1,\ldots,B\}
\big\}.
\label{eqn:PSet}
\end{align}

\subsubsection{Power Allocation as Constrained Optimization}
\label{sec:constrained_opt}
We assess power allocation based on the throughput supported for each communication framework. To formulate this objective, first, we note that for orthogonal communication over the \ac{awgn} channel in \eqref{eq:received message b}, the achievable rate for communicating the $k$th message over frequency band $b$ and link $(i\!\to\!j)\in\mySet{E}$  is given by
\begin{equation}
R_{i\to j,k}^{(b)}(\myMat{P})
=
\log_2\left(
1+
\frac{
\left|h_{i\to j}^{(b)}\right|^2
\left[\myMat{P}\right]_{b,k,i,j}^{2}
}{
\sigma_b^2
+
I_{i\to j}^{(b)}(\myMat{P})
}
\right),
\label{eq:link_rate}
\end{equation}
where
\[
I_{i\rightarrow j}^{(b)}(\myMat{P})
=
\sum_{l\in\mySet{N}_j}
\left|h_{l\rightarrow j}^{(b)}\right|^{2}
\sum_{k=1}^{K}
\sum_{q\in\mySet{N}_l}
P_{b,k,l,q}^{2}.
\]
Following the widely adopted {\em treating-interference-as-noise} assumption, co-channel interference from neighboring transmitters is modeled as additive Gaussian noise at the receiver~\cite{geng2015optimality}. The interference model assumes that each neighboring transmitter occupies a single outgoing transmission on a given communication resource. This is consistent with the per-link message allocation constraint introduced in~\eqref{eqn:PSet}.

When communicating over multiple links, the throughput is dictated by the weakest link. Accordingly, for a given connected subgraph of $\mySet{G}$ with edges $\psi \subseteq \mySet{E}$ satisfying the communication requirements of the considered framework, the rate one can achieve for conveying the $k$th message over channel $b$ is 
 \begin{equation}
    R_{\psi,k}^{(b)}(\myMat{P}) = \min_{(i\to j)\in\psi} R_{i\to j,k}^{(b)}(\myMat{P}).
    \label{eq:subgraph_rate}
\end{equation}

In multi-channel \acp{manet}, a transmitter can encode a message over several different channels, and possibly employ different routes over different channels for the same message. Therefore, a transmitter $u^{\rm Tx}$ can convey its $k$th message to a set of $Q$ intended receivers $\{u^{\rm Rx}_1,\ldots,u^{\rm Rx}_Q\}$ by selecting the route with maximal rate for each channel, achieving end-to-end rate of
\begin{equation}
    R_k^{\rm E2E} (\myMat{P}; \Phi)
    = \sum_{b=1}^{B} \max_{\psi\in\Phi} R_{\psi,k}^{(b)}(\mathbf{P}),
    \label{eq:Rk_mc}
\end{equation}
where $\Phi$ denotes the set of all subgraphs of $\mySet{G}$ containing $u^{\rm Tx}$ and $\{u^{\rm Rx}_1,\ldots,u^{\rm Rx}_Q\}$.  
A power allocation $\myMat{P}$ is assessed by the end-to-end rate that can be achieved for all messages; namely, the power allocation problem can be written as 
\begin{equation} 
    \myMat{P}^\star
    = \argmax_{\myMat{P}\in\mySet{P}}  \min_{k\in \{1,\ldots,K\}}
      R_k^{\rm E2E} (\myMat{P}; \Phi_k). 
    \label{eqn: best power allocation_fw}
\end{equation}

The unified problem formulation in \eqref{eqn: best power allocation_fw} accommodates \ref{item:unicast}-\ref{item:many2many}, which vary in the parameters of the optimization. For \ref{item:unicast}-\ref{item:multicast}, there is only one message (i.e., $K=1$), with $\Phi$ specializing into all paths from $u^{\rm Tx}$ to the single receiver $u^{\rm Rx}$ for \ref{item:unicast}, and including all subgraphs including all $Q$ intended receivers for \ref{item:multicast}. 
The frameworks with $K>1$ messages differ in the setting of the set of subgraphs $\Phi_k$, which for \ref{item:multicom} accommodates the set of paths from the single transmitter $u^{\rm Tx}$ to $u^{\rm Rx}_k$, for \ref{item:convergecast} dictates all paths from $u^{\rm Tx}_k$ to the single receiver $u^{\rm Rx}$,and for \ref{item:many2many} contains all paths from $u^{\rm Tx}_k$ to $u^{\rm Rx}_k$.

The candidate set $\Phi_k$ is constructed offline according to the communication framework. For \ref{item:unicast} and \ref{item:multicom}-\ref{item:many2many}, it contains all simple source--destination paths, found by depth-first search without cycles. For \ref{item:multicast}, $\Phi$ contains all connected subgraphs spanning the source and all intended receivers. These sets are used only during offline training for objective evaluation and gradient computation; at inference, MANET-GNN outputs the power allocation after a fixed number of message-passing rounds, without path or subgraph enumeration.

\subsubsection{Problem Formulation}
\label{sec:Prob Formulation}

Our goal is to design a policy for setting the power allocations $\myMat{P}$ based on the centralized optimization objective in \eqref{eqn: best power allocation_fw}, under the different frameworks \ref{item:unicast}-\ref{item:many2many}. 
While \eqref{eqn: best power allocation_fw} is formulated as a centralized optimization setting, i.e., a mapping of the full \ac{manet} \ac{csi} $\{h_{i \to j}^{(b)}\}$ into $\myMat{P}^{\star}_{\rm fw}$, we aim to design a method that meets the following requirements: 
\begin{enumerate}[label={R\arabic*}]
    \item \label{itm:Decent} \emph{Decentralized operation}, i.e., each node $i$ sets its own  $\{p_{i \to j}^{(b)}\}$ (and, when applicable, $\{p_{i \to j,k}^{(b)}\}$) based on knowing its neighbors ($\mySet{N}_i$) and its local \ac{csi} $\{\hat{h}_{i \to j}^{(b)}\}_{j\in \mySet{N}_i,\, b \in \{1,\ldots B\}}$. 
    \item  \label{itm:Messages} \emph{Limited latency optimization}, where each node $i$ is allowed to exchange at most $L$ messages with its neighbors $\mySet{N}_i$. 
    \item \label{itm:transfer} The method should be applicable on \emph{different topologies}, i.e., generalize across graphs with varying sizes and topologies. 
    \item \label{itm:Noisy} The local \ac{csi}  $\{\hat{h}_{i \to j}^{(b)}\}_{j\in \mySet{N}_i,\, b \in \{1,\ldots B\}}$ may be a \emph{noisy estimate} of the actual \ac{csi}.  
\end{enumerate}

To cope with \ref{itm:Decent}--\ref{itm:Noisy}, we assume access during design to \ac{csi} from various \ac{manet} realizations, represented by the data set
\begin{equation}
    \mySet{D} = \left\{\{h_{i \to j, d}^{(b)}\}_{(i,j)\in \mySet{E}_d,\, b \in \{1,\ldots B\}},\ \mySet{G}_d = (\mySet{V}_d, \mySet{E}_d)\right\}_{d=1}^{|\mySet{D}|},
    \label{eqn:DataSet}
\end{equation}
with $h_{i \to j, d}^{(b)}$ representing the realization of $h_{i \to j}^{(b)}$ in the $d$th \ac{manet} in $\mySet{D}$. 
Note that \eqref{eqn:DataSet} does not contain ground-truth power allocations, and that its \acp{csi} come from different \ac{manet}s with different topologies and channel conditions.

\vspace{-0.1cm}
\section{Decentralized Learned Optimization}
\label{sec:Decentralized}

In this section, we present our decentralized learned optimization framework for the power allocation problem described in Subsection~\ref{sec:Problem}. 
We leverage the ability to formulate the allocation problem globally for all considered communication frameworks, with the objective \eqref{eqn: best power allocation_fw} depending on end-to-end paths or subgraphs and on per-node power constraints. 
To address requirements \ref{itm:Decent}--\ref{itm:Noisy}, we propose to tune the power allocation policy by leveraging the optimization formulation in \eqref{eqn: best power allocation_fw} through \emph{learned optimization} tools. 
Our design builds upon the empirical success of \acp{gnn} in solving learned optimization tasks~\cite{shen2020graph, randall2022grows}, exploiting their inherent ability to operate in a decentralized manner (\ref{itm:Decent}) while naturally adapting to different network topologies (\ref{itm:transfer})~\cite{corso2024graph}. 

Specifically, we introduce a dedicated \ac{gnn} architecture, termed \emph{MANET-GNN}. The architecture of MANET-GNN, detailed in Subsection~\ref{sec:arch}, is inspired by message-passing networks~\cite{feng2022powerful}, and explicitly constrains the number of message exchanges to meet the latency requirement (\ref{itm:Messages}), while being inherently scalable to different topologies (\ref{itm:transfer}).  
To enable efficient decentralized tuning of the power $\myMat{P}$, we propose a dedicated training method in Subsection~\ref{sec:training}, that trains MANET-GNN as a \emph{distributed learned optimizer}. Our learning formulation encourages MANET-GNN  to approximate the centralized power allocation solutions in a manner accommodating \ref{item:unicast}-\ref{item:many2many}, while leveraging the casting of the \ac{gnn} as an optimizer to learn in an unsupervised manner, i.e., without requiring ground-truth labels. We conclude with a discussion in Subsection~\ref{sec:discussion}.
\vspace{-0.1cm}
\subsection{MANET-GNN Distributed Optimizer Architecture}
\label{sec:arch}

MANET-GNN consists of a gated message-passing backbone followed by shared per-edge decoders that map the learned node and edge embeddings into framework-specific power and routing variables. The architecture encodes the local \ac{csi}, topology, and communication roles into graph signals, processes them through a fixed number of decentralized message-passing rounds, and outputs a feasible power allocation.

\subsubsection{Input Encoding}

As formulated in Subsection~\ref{sec:Problem}, the allocation depends on the \ac{manet} topology, the available \ac{csi} $\{h_{i\to j}^{(b)}\}_{b=1}^{B}$, and the communication framework. MANET-GNN represents these quantities as multivariate node and edge features over the communication graph~\cite{ortega2018graph}.

For each edge $(i,j)\in\mySet{E}$, the input edge feature stacks the real and imaginary parts of the channel gains:
\begin{equation*}
 \myVec{e}_{i \rightarrow j}^{(0)}
 =
 \big[
 {\rm Re}\{h_{i\to j}^{(1)},\ldots,h_{i\to j}^{(B)}\}
 \,\|\,
 {\rm Im}\{h_{i\to j}^{(1)},\ldots,h_{i\to j}^{(B)}\}
 \big]^\top .
\end{equation*}

The node feature combines a coarse equal-split power prior with a role encoding. The role vector $\myVec{r}_i$ identifies whether node $i$ acts as a transmitter, receiver, relay, and as part of a specific commodity. Using the equal-split initialization
$p_{i\to j}^{(b)} = 1/\sqrt{|\mySet{N}_i|B}$, the input node feature is
\begin{equation}
    \myVec{x}_i^{(0)}
    =
    \Big[
    \sum_{j\in\mySet{N}_i}
    \big[
    p_{i\to j}^{(1)},\ldots,p_{i\to j}^{(B)},
    p_{j\to i}^{(1)},\ldots,p_{j\to i}^{(B)}
    \big]
    \,\|\, \myVec{r}_i^\top
    \Big]^\top .
    \label{eqn:NodeEmb1}
\end{equation}
Only the first $B$ entries are used as the initial node embedding. 

\subsubsection{Gated Message-Passing Backbone}

The MANET-GNN backbone stacks $L/2$ gated \ac{gnn} layers. Each layer corresponds to two neighbor-to-neighbor message exchanges and updates the node and edge embeddings locally. At layer $l$, node $i$ uses its embedding $\myVec{x}_i^{(l-1)}$, its incident edge embeddings, and messages received from neighboring nodes.

For each edge $(i\to j)$, the normalized node and edge embeddings are used to compute an edge update:
\begin{subequations}
\label{eqn:edgeFeat}
\begin{align}
\Delta \myVec{e}_{i\to j}^{(l)}
&=
\mathrm{MLP}_{\rm e}^{(l)}
\big(
\bar{\myVec{e}}_{i\to j}^{(l-1)}
\,\|\,
\bar{\myVec{x}}_j^{(l-1)}
\,\|\,
\bar{\myVec{x}}_i^{(l-1)}
\big), \\
\myVec{e}_{i\to j}^{(l)}
&=
\mathrm{LN}
\Big(
\myVec{e}_{i\to j}^{(l-1)}
+
\sigma(\Delta \myVec{e}_{i\to j}^{(l)})
\odot
\Delta \myVec{e}_{i\to j}^{(l)}
\Big),
\end{align}
\end{subequations}
where $\mathrm{LN}(\cdot)$ is layer normalization and $\sigma(\cdot)$ is  element-wise sigmoid. The gate adaptively scales each feature of the edge update before the residual addition based learned importance.

The refined edge embedding parametrizes a \ac{film}~\cite{brockschmidt2020gnn} modulation of the message:
\begin{equation}
\myVec{m}_{i\to j}^{(l)}
=
\big(\mathbf{1}+\myVec{\gamma}_{i\to j}^{(l)}\big)
\odot
\myMat{W}_{\rm msg}^{(l)}
\bar{\myVec{x}}_i^{(l-1)}
+
\myVec{\beta}_{i\to j}^{(l)}
\in\mathbb{R}^{B},
\label{eqn:MessageGen}
\end{equation}
with
$\myVec{\gamma}_{i\to j}^{(l)}=\myMat{W}_{\gamma}^{(l)}\myVec{e}_{i\to j}^{(l)}$
and
$\myVec{\beta}_{i\to j}^{(l)}=\myMat{W}_{\beta}^{(l)}\myVec{e}_{i\to j}^{(l)}$, where
$
\mathbf{W}_{\mathrm{msg}}^{(l)}
\in
\mathbb{R}^{D_n^{(l+1)}\times D_n^{(l)}},
\qquad
\mathbf{W}_{\gamma}^{(l)}, \mathbf{W}_{\beta}^{(l)}
\in
\mathbb{R}^{D_n^{(l+1)}\times D_e^{(l+1)}}.
$
Here, $D_n^{(l)}$ and $D_e^{(l)}$ denote the node and edge embedding dimensions at layer $l$, respectively.
All hidden node and edge embeddings have dimension $B$.
Node $i$ then aggregates the incoming messages and updates its embedding via
\begin{equation}
    \myVec{x}_i^{(l)}
    =
    \mathrm{LN}
    \left(
    \myVec{x}_{i}^{(l-1)}
    +
    \mathrm{MLP}_{\rm a}^{(l)}
    \left(
    \frac{1}{|\mySet{N}_i|}
    \sum_{j\in\mySet{N}_i}
    \myVec{m}_{j\to i}^{(l)}
    \right)
    \right).
    \label{eqn:NodeUpdate}
\end{equation}
In the first layer, if $\myVec{x}_{i}^{(0)}$ has more than $B$ entries, only its first $B$ entries are used in the residual connection.

\begin{figure*}
    \centering
    \begin{subfigure}{0.67\textwidth}
        \centering
        \includegraphics[width=\textwidth]{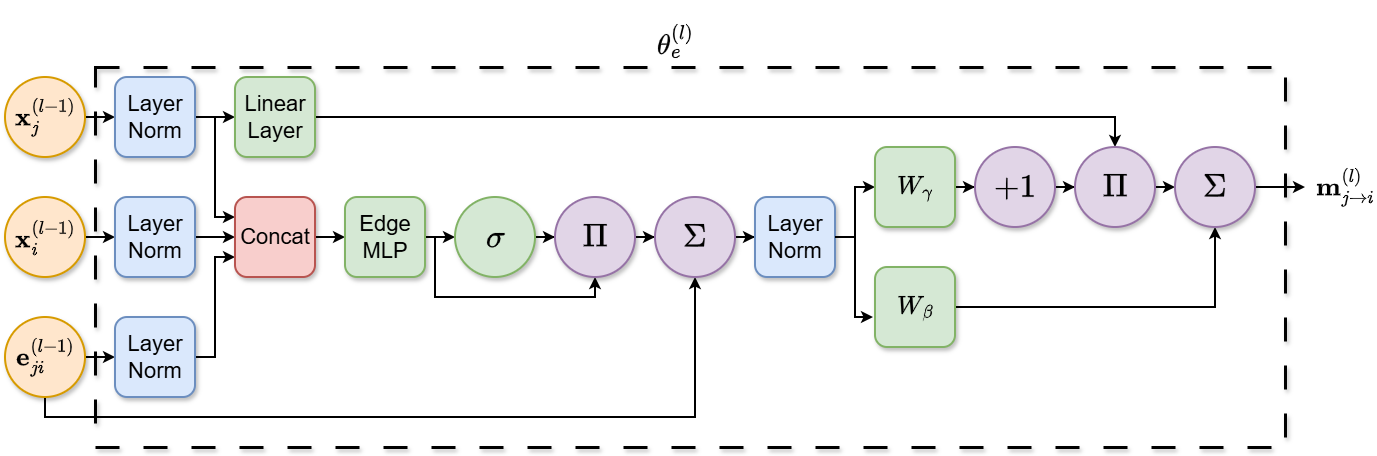}
        \caption{Encoder}
        \label{fig:Encoder}
    \end{subfigure}%
    \begin{subfigure}{0.22\textwidth}
        \centering
        \includegraphics[height=0.12\textheight]{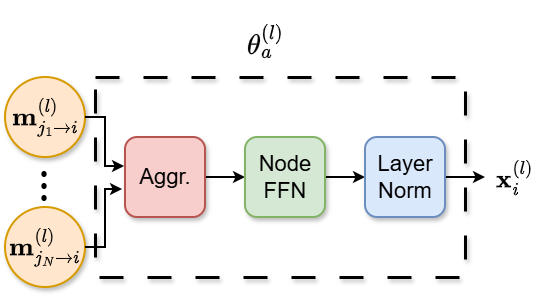}
        \caption{Aggregator}
        \label{fig:Aggregator}
    \end{subfigure}
    \caption{Gated message-passing layer of MANET-GNN.}
    \label{fig:gated_layer}
    \vspace{-0.2cm}
    \label{fig:manetgnn}
\end{figure*}

\subsubsection{Output Processing}

After $L/2$ gated layers, each node applies shared decoders to its outgoing links. For link $(i\to j)$, the decoder input is
\begin{equation}
   \myVec{f}_{i\to j}
    =
    \big[
    \myVec{e}_{i\to j}^{(L/2)}
    \,\|\,
    \myVec{x}_i^{(L/2)}
    \,\|\,
    \myVec{x}_j^{(L/2)}
    \big].
    \label{eqn:FeaturesEnd}
\end{equation}

\begin{enumerate}
    \item {\em Tentative power allocations}: Denoted $ \tilde{\myVec{p}}{i\to j} \in \mathbb{R}{+}^B$, obtained by applying a \ac{fc} layer with Softplus activation to $\myVec{f}_{i\to j}$.
    \item{\em Candidate-Based Routing Head}: 
    For the multi-message frameworks \ref{item:multicom}-\ref{item:many2many}, routing variables must represent complete routes rather than independent link activations. For commodity $k$, with source $s_k$ and destination $d_k$, we define the local candidate set
    \begin{equation*}
     \hspace{-0.2cm}   \mathcal{P}_k^{(L/2)}
        =
        \left\{
        p_{k,m}: p_{k,m}\text{ connects }s_k\text{ to }d_k,\;
        |p_{k,m}|\leq L/2
        \right\}_{m=1}^{M_k}.
    \end{equation*}
    The same $L/2$-hop candidate set is used during training and inference. Thus, route selection requires only locally available information and does not rely on global topology knowledge or a centralized route table.

    For each candidate path, the routing head forms a path embedding
   $
        \bar{\myVec{e}}_{k,m}^{(l)}
        =
        \frac{1}{|p_{k,m}|}
        \sum_{(i,j)\in p_{k,m}}
        \myVec{e}_{i\to j}^{(l)}$,
    that is concatenated with band-dependent  route features
    \begin{equation}
        \myVec{r}_{b,k,m}
        =
        \left[
        g^{\min}_{b,k,m},
        g^{\mathrm{avg}}_{b,k,m},
        |p_{k,m}|,
        \eta_{b,k,m}
        \right],
    \end{equation}
    where $g^{\min}_{b,k,m}
        =
        \min_{(i,j)\in p_{k,m}}
        |h_{i\to j}^{(b)}|^2$,  
        $g^{\mathrm{avg}}_{b,k,m}
        =
        \frac{1}{|p_{k,m}|}
        \sum_{(i,j)\in p_{k,m}}
        |h_{i\to j}^{(b)}|^2$,  and
       $\eta_{b,k,m}
        =
        \sum_{(i,j)\in p_{k,m}}
    \sum_{q\in\mySet{N}_j^{(L/2)}\setminus\{i\}}
        |h_{q\to j}^{(b)}|^2$. 
    The last term measures local receiver-side interference exposure.

    A shared \ac{mlp} assigns a band-dependent score to each candidate:
    \begin{equation}
        a_{b,k,m}
        =
        \phi_{\mathrm{route}}
        \left(
        \left[
        \bar{\myVec{e}}_{k,m}^{(l)},
        \myVec{r}_{b,k,m}
        \right]
        \right).
    \end{equation}
    During training, differentiability is preserved by applying a soft selection over the candidate paths of each band:
    \begin{equation}
        \pi_{b,k,m}
        =
        \frac{\exp(a_{b,k,m}/\tau_r)}
        {\sum_{m'=1}^{M_k}\exp(a_{b,k,m'}/\tau_r)} .
    \end{equation}
    The routing tensor is constructed by projecting the selected candidates onto their constituent edges:
    \begin{equation}
        Z_{b,k,i,j}
        =
        \sum_{m=1}^{M_k}
        \pi_{b,k,m}
        \mathbbm{1}\{(i,j)\in p_{k,m}\}.
        \label{eq:candidate_z_construction}
    \end{equation}
    Hence, nonzero entries of $Z_{b,k,:,:}$ represent candidate routes.

    At inference, the soft selection is replaced by a hard per-band decision:
    \begin{equation}
        m^\star_{b,k}
        =
        \argmax_{m:\,p_{k,m}\in\mathcal{P}_k^{(L/2)}}
        a_{b,k,m}.
    \end{equation}
    Then $Z_{b,k,i,j}=1$ only for edges belonging to $p_{k,m^\star_{b,k}}$. Since the maximization is performed independently for each band, a commodity may use different routes over different frequency bands, consistently with \eqref{eq:Rk_mc}.

    The routing and power-allocation heads are trained jointly using the full differentiable objective, including the interference generated by all commodities. Restricting training and inference to the same local candidate sets avoids a train--test mismatch, enforces route consistency by construction, and preserves decentralized multi-band routing flexibility.
\end{enumerate}

The final transmission amplitude assigned by node $i$ to message $k$ over link $(i,j)$ is obtained by masking the raw amplitude with the routing variable and normalizing it to satisfy the per-node power constraint:
\begin{equation}
\label{eqn:Peff}
    [\myMat{P}]_{:,k,i,j}
    =
    \frac{
    \sqrt{[\myVec{z}_{i\to j,k}]}\odot \tilde{\myVec{p}}_{i\to j,k}
    }
    {
    \sqrt{
    \sum_{k'=1}^{K}
    \sum_{j'\in\mySet{N}_i}
    \left\|
    \sqrt{[\myVec{z}_{i\to j',k'}]}
    \odot
    \tilde{\myVec{p}}_{i\to j',k'}
    \right\|_2^2
    }
    } .
\end{equation}

\subsubsection{Overall Algorithm}

MANET-GNN operates as a learned decentralized optimizer. Each gated layer implements two neighbor-to-neighbor message exchanges, and the depth $L/2$ directly controls the number of communication rounds and the receptive field. After $L/2$ rounds, each node has access only to information within its $L/2$-hop neighborhood; dependencies beyond this range cannot be captured in a single forward pass. Increasing the depth therefore enables longer-range coordination, at the cost of additional communication and computation.

The trainable parameters of MANET-GNN are
\begin{equation}
    \myVec{\theta}
    =
    \Big\{
    \{\myVec{\theta}_{\rm e}^{(l)}, \myVec{\theta}_{\rm a}^{(l)}\}_{l=1}^{L/2},
    \myVec{\theta}_{\rm d}^{(P)},
    \myVec{\theta}_{\rm d}^{(Z)}
    \Big\}.
    \label{eqn:parameters}
\end{equation}

During inference, nodes execute only the forward pass of MANET-GNN, yielding a fully decentralized power-allocation policy. The overall flow is illustrated in Fig.~\ref{fig:manetgnn}, where the stacked gated layers serve as iterations of a learned optimizer and the final decoders output the routing-aware power allocation.
\begin{algorithm}[t]
\caption{MANET-GNN  at Node $i$ (Layer $l$)}
\label{alg:manet_gnn_node}
\SetKwInOut{Initialization}{Init}
\Initialization{Trained   $\myVec{\theta}$; neighbors $\mySet{N}_i$} 
\SetKwInOut{Input}{Input}
\Input{
Node features $\myVec{x}_i^{(l-1)}$; neighbors node features $\{\myVec{x}_j^{(l-1)}\}_{j\in\mySet{N}_i}$; edge features $\{\myVec{e}_{i\to j}^{(l-1)}\}_{j\in\mySet{N}_i}$; 
}

\nonl\underline{\textbf{Message encode:}}\\
\For{$j\in\mySet{N}_i$}{
Update edge embedding $\myVec{e}_{i\to j}^{(l)}$ via \eqref{eqn:edgeFeat}\; 
Compute message  $\myVec{m}_{i\to j}^{(l)}$ via \eqref{eqn:MessageGen}\;
Send  $\myVec{m}_{i\to j}^{(l)}$ to node $j$ \tcp*{Message pass 1}
}
\nonl\underline{\textbf{Aggregate:}}\\ 
Update $\myVec{x}_i^{(l)}$ via \eqref{eqn:NodeUpdate}\;
Broadcast  $\myVec{x}_{i}^{(l)}$ to neighbors \tcp*{Message pass 2}
\nonl\underline{\textbf{Power setting:}}\\
\If{Last layer}
{\For{$j\in\mySet{N}_i$}{
Compute features $\myVec{f}_{i\rightarrow j}$ via \eqref{eqn:FeaturesEnd}\;
Compute $\tilde{\myVec{p}}_{i\to j} $ and ${\myVec{z}}_{i\to j} $ via \ac{fc} network and routing head\;
}
Set power allocation $[\myMat{P}]_{:,:,i,:}$ via \eqref{eqn:Peff}\;
}
\end{algorithm}

\vspace{-0.1cm}
\subsection{Training Across Communication Frameworks}
\label{sec:training}

The parameters of MANET-GNN \eqref{eqn:parameters} are trained in an unsupervised fashion using the unlabeled dataset $\mySet{D}$ in \eqref{eqn:DataSet}. 
The key ideas underlying our training procedure are: $(i)$  treat MANET-GNN as a parameterized mapping from \ac{csi} and topology to power allocations; $(ii)$ formulate a loss that  encourages it to gradually improve the objective $R_k^{\rm E2E}$, without requiring ground-truth power labels; and $(iii)$ account for the expected presence of noisy \ac{csi} by injecting noise in training.

\subsubsection{Training Loss}
Given a sample $(\mySet{G}_d,\{h_{i\to j,d}^{(b)}\})$ from $\mySet{D}$, we denote the sequence of power allocations produced by processing the output of the $l$th layer MANET-GNN as $\myMat{P}^{(l)}(\mySet{G}_d, \{h_{i\to j,d}^{(b)}\};\myVec{\theta})$. 
The rate expressions used in the loss are evaluated according to the \ac{sinr}-based formulation of Section \ref{sec:System}, and therefore account for both channel noise and co-channel interference.
Based on the optimization objective in \eqref{eqn: best power allocation_fw}, we compute the rate achieved as
\begin{equation}
  R_d^{(l)}(\myVec{\theta})
  \triangleq \min_{k\in \{1,\ldots,K\}}
      R_k^{\rm E2E} (\myMat{P}^{(l)}(\mySet{G}_d, \{h_{i\to j,d}^{(b)}\};\myVec{\theta}); \Phi_k).
      \label{eqn:rateEv}
\end{equation}
The primary empirical loss term is set to 
\begin{equation}
   \mathcal{L}_{\mySet{D}}^{\rm rate}(\myVec{\theta})
   = -\frac{1}{|\mySet{D}|} \sum_{d=1}^{|\mySet{D}|} R_d^{(L/2)}(\myVec{\theta}),
\end{equation}
which encourages the final layer to maximize this rate.

The objective in \eqref{eqn: best power allocation_fw} contains a minimum over the links of each route and a maximum over the feasible routes. These operations capture the route bottleneck and route selection, respectively, but are non-smooth and therefore unsuitable for gradient-based training. We replace them with differentiable soft approximations.

For a path $\psi$, the bottleneck rate on band $b$ for message $k$ is approximated as
\begin{equation*}
\widetilde{R}_{\psi,k}^{(b)}
=
-\frac{1}{\tau_{\min}}
\log
\sum_{(i,j)\in\psi}
\exp\left(
-\tau_{\min}R_{i\to j,k}^{(b)}
\right),
\end{equation*}
where $\tau_{\min}>0$. Route selection is similarly approximated by
\begin{equation*}
\widetilde{R}_{k}^{(b)}
=
\frac{1}{\tau_{\max}}
\log
\sum_{\psi\in\Phi_k}
\exp\left(
\tau_{\max}\widetilde{R}_{\psi,k}^{(b)}
\right),
\end{equation*}
where $\tau_{\max}>0$. Increasing $\tau_{\min}$ and $\tau_{\max}$ sharpens the approximations toward the corresponding minimum and maximum operators. These relaxations are used only for training; inference and evaluation employ feasible allocations and the original communication objective.
The candidate path/subgraph sets $\Phi_k$ required to evaluate the training objective are constructed offline. Their preprocessing cost increases with the network size, connectivity density, and number of commodities. This global search is not part of the MANET-GNN forward pass: during deployment, the nodes produce the allocation from locally available \ac{csi} and exchanged messages.

To stabilize training and to better align MANET-GNN with the notion of a learned iterative optimizer, we impose a monotonicity regularizer that encourages the rate to improve across consecutive layers \cite{shlezinger2025deep}. 
For each sample we measure the rate differences $\Delta R_d^{(l)}(\myVec{\theta}) = R_d^{(l+1)}(\myVec{\theta}) - R_d^{(l)}(\myVec{\theta})$, and penalize violations of a small positive margin $\delta>0$ via
\begin{equation}
\mathcal{L}_{\mySet{D}}^{\rm mono}(\myVec{\theta}) = \frac{1}{|\mySet{D}|L} \sum_{d=1}^{|\mySet{D}|}\sum_{l=1}^{L/2-1} \max\big( \delta - \Delta R_d^{(l)}(\myVec{\theta}),0 \big).
\end{equation}
To encourage route concentration in the single-message frameworks \ref{item:unicast}-\ref{item:multicast}, we add a sparsity penalty on the transmission amplitudes. 
Let: $a_{i,j}=\sqrt{\sum_{b=1}^{B} P_{b,i,j}^{2}}$
denote the aggregate amplitude assigned to edge $(i,j)$. The penalty is defined as
\begin{equation}
    \label{sparsity loss}
    \mathcal{L}_{\mySet{D}}^{\rm sparse}(\myVec{\theta})
    =
    |R_{\mySet{D}}(\myVec{\theta})|
    \left(
    \frac{\|\myMat{A}\|_1}{\|\myMat{A}\|_2}
    -
  ,  1
    \right),
\end{equation}

where $\myMat{A}=[a_{i,j}]$ and $R$ is the achieved rate. The ratio $\frac{\|\myMat{A}\|_1}{\|\myMat{A}\|_2}$ is minimized when the power is concentrated on fewer edges,  encouraging  to form a compact route from the source to the destination(s).
The overall loss takes the form
\begin{equation}
\label{eq: train loss}
   \mathcal{L}_{\mySet{D}}(\myVec{\theta})
   = \mathcal{L}_{\mySet{D}}^{\rm rate}(\myVec{\theta})
   + \lambda_m\, \mathcal{L}_{\mySet{D}}^{\rm mono}(\myVec{\theta})
   +\lambda_s\mathcal{L}^{\rm sparse}_{\mySet{D}}(\myVec{\theta}),
\end{equation}
with $\lambda_m\geq 0$ controlling the strength of the monotonicity regularization, and $\lambda_{\mathrm{s}}\geq 0$ is the sparsity weight. 
This loss is minimized by mini-batch \ac{sgd} type learning.

\subsubsection{Noisy-CSI-Aware Training}
To enhance robustness to imperfect \ac{csi} and satisfy requirement \ref{itm:Noisy}, we train MANET-GNN under a noisy-\ac{csi} regime. 
In particular, for each sample in a batch we replace the true channel realization $\{h_{i\to j,d}^{(b)}\}$ used in the forward pass with an estimated version $\{\hat{h}_{i\to j,d}^{(b)}\}$, obtained using a \ac{lmmse} channel estimator under the assumed \ac{awgn} observation model. The model thus learns to produce power allocations based on estimates that mimic those available at run time.

While the input \ac{csi} provided to MANET-GNN in training is noisy,  the loss $\mathcal{L}_{\mySet{D}}(\myVec{\theta})$ is always evaluated with respect to the true \ac{csi}, so that improvements in the learned policy are measured in terms of the actual end-to-end performance under perfect knowledge. 
This training protocol can be interpreted as an adversarial learned optimization scheme~\cite{sofer2025unveiling}, in which the optimizer is trained to be robust to perturbations in its inputs.
The overall training procedure based on \ac{sgd} is summarized as Algorithm~\ref{alg:train}.

\begin{algorithm}[t]
\caption{MANET-GNN \ac{sgd} Training}
\label{alg:train}
\SetKwInOut{Initialization}{Init}
\Initialization{
Initial parameters $\myVec{\theta}$; Learning rate $\eta$; \#epochs ${\rm epoch}_{\max}$; \#batches $Q$; Hyperparameters $\lambda,\delta, L$; 
}
\SetKwInOut{Input}{Input}
\Input{
Training set $\mySet{D}=\{(\mySet{G}_d,\{\hat{h}_{i\rightarrow j, d}^{(b)}\}\}_{d=1}^{|\mySet{D}|}$; \\Framework Tx/Rx sets; \#Commodities $K$. 
}

\For{${\rm epoch}=0,1,\ldots,{\rm epoch}_{\max}-1$}{
    Randomly divide $\mySet{D}$ into $Q$ batches $\{\mySet{D}_q\}_{q=1}^{Q}$\;
    \For{$q=1,\ldots,Q$}{
        Apply \ac{manet} \ac{gnn} $\myVec{\theta}$ to $\{\mySet{G}_d,\{\hat{h}_{i\rightarrow j, d}^{(b)}\}\}_{d\in \mySet{D}_q}$\;
        Compute loss 
        $\mathcal{L}_{\mySet{D}_q}(\myVec{\theta})$ via~\eqref{eq: train loss}\;  
        Update $\myVec{\theta}\leftarrow \myVec{\theta}-\eta\nabla_{\myVec{\theta}}\mathcal{L}_{\mySet{D}_q}(\myVec{\theta})$\;
    }
}
\KwRet{$\myVec{\theta}$}
\end{algorithm}

\subsection{Discussion}
\label{sec:discussion}

The proposed MANET-GNN framework provides a principled mechanism for meeting the requirements \ref{itm:Decent}--\ref{itm:Noisy} across all considered communication frameworks. 
First, its decentralized message-passing design inherently ensures that each node sets its local power allocations based solely on its local \ac{csi} and messages from its neighbors, thereby satisfying \ref{itm:Decent}. 
The explicit limitation on the number of message-passing rounds directly controls the number of communication exchanges between neighboring nodes and thus provides a handle on latency and overhead, addressing \ref{itm:Messages}. 
The use of \acp{gnn} with shared parameters across nodes and edges allows the same model to operate on graphs with different sizes and topologies, enabling generalization across heterogeneous \acp{manet} as required by \ref{itm:transfer}. 
Finally, the noisy-\ac{csi}-aware training procedure equips the model with robustness to estimation errors and mismatches between training and deployment conditions, thereby tackling \ref{itm:Noisy}. 
Because unsupported communication resources are represented
through zero-valued channel coefficients, the resulting topology
is naturally encoded in the graph representation. Therefore,
the proposed MANET-GNN architecture can operate on heterogeneous
networks without architectural modifications.
 
While our framework and the architecture of MANET-GNN accommodates multiple different communication frameworks \ref{item:unicast}-\ref{item:many2many}, the formulation of the loss in \eqref{eq: train loss} is framework-dependent by \eqref{eqn:rateEv}. Accordingly, the specific weights of MANET-GNN are trained for a given framework, and modularity can be supported by training a set of MANET-GNN weight configurations, one per communication framework, while fixing a maximum supported number of commodities $K$ at the firmware level. One can potentially extend this approach into a single multi-framework model via, e.g., hypernetworks~\cite{raviv2025modular} or mixture-of-experts formulations~\cite{masoudnia2014mixture}. We leave these extensions for future work.

We consider joint routing and power allocation under fixed channel assignments: links may reuse channels and cause co-channel interference, but each message remains on a single channel along its multi-hop route within a transmission block. Per-hop channel switching would require additional channel-assignment variables and is left for future work.
Future work may extend the framework to decentralized OFDMA, where routing, power, and band assignment are jointly optimized, as well as to heterogeneous per-flow QoS requirements through weighted utilities. Finally, the present work considers single-antenna communications for each channel. Extending the proposed decentralized learned optimization framework to multi-antenna systems by jointly optimizing beamforming and spatial resource allocation is an important direction for future research.

\vspace{-0.1cm}
\section{Experimental Study}
\label{sec: Experimental}
Here, we numerically evaluate MANET-GNN\footnote{The source code used in our empirical study, along with the hyperparameters is available at \url{https://github.com/AlterTomer/Decentralized-MANET}}, with the aims of:  
$(i)$ validate that it correctly instantiates classical communication paradigms of \ref{item:unicast}-\ref{item:many2many} from the formulation of the unified parent problem;  
$(ii)$  assess performance relative to centralized and heuristic baselines under varying network sizes, channel conditions, and levels of \ac{csi}; and  
$(iii)$ examine scalability, robustness, and inference latency in representative \ac{manet} settings.  

\vspace{-0.1cm}
\subsection{Setup}
\subsubsection{\ac{manet} Generation}
\label{sec:graph_generation}
We generate \ac{manet} topologies by sampling an undirected random graph over \( |\mathcal{V}| \) nodes. The adjacency matrix \( \mathbf{A} \in \{0,1\}^{|\mathcal{V}|\times|\mathcal{V}|} \) is constructed by drawing each off-diagonal entry independently from a Bernoulli distribution with edge probability $p$, followed by symmetrization to enforce undirected connectivity while avoiding isolated nodes. 
During training and testing, the datasets included graphs with $p\in\{0.1, 0.2, 0.3, 0.4, 0.5\}$.

\subsubsection{Channel Generation}
\label{sec:channel_generation}

 For each  realization, we generate complex reciprocal channels $\{h_{i\rightarrow j}^{(b)}\}$ over feasible links, i.e., \(h_{i\rightarrow j}^{(b)}=0\) whenever \(\mathbf{A}_{ij}=0\) and $h_{i\rightarrow j}^{(b)} = h_{j\rightarrow i}^{(b)}$.
For multi-message settings, the physical channel realization is shared across all message indices \(k \in \{1,2,\dots, K\}\), i.e., the \ac{manet} physical channel  is independent of the communication framework. 

We consider {\em QuaDRiGa} frequency-selective channels~\cite{Jaeckel2014QuaDRiGa}. Nodes are placed in a \(100\times 100\)~m urban block with heights drawn such that most nodes are at street level and a minority are elevated (second-floor) users.  For every active edge \(\{i,j\}\in\mathcal{E}\), we instantiate an omnidirectional link in QuaDRiGa fixed urban scenario using the normalized  channel over \(B\) sub-channels.
The dynamic \ac{snr} range is controlled across experiments by varying the noise level $\sigma_b^2$, while keeping channel statistics fixed, defining ${\rm SNR}^{(b)} = 10\cdot \log_{10}\big(\frac{1}{\sigma_b^2}\big)$.

We evaluate our framework under two \ac{csi} regimes: full \ac{csi} and noisy \ac{csi}. This allows us to disentangle the performance of the learned decentralized optimizer from the impact of channel estimation errors.
In the full CSI setting, each node is assumed to have access to the true channel coefficients. In the noisy \ac{csi} setting, \ac{csi} is obtained through pilot-based estimates.

\begin{figure*}[!tbp]
    \centering
    \begin{subfigure}[t]{0.43\textwidth}
        \centering
        \includegraphics[width=\linewidth]{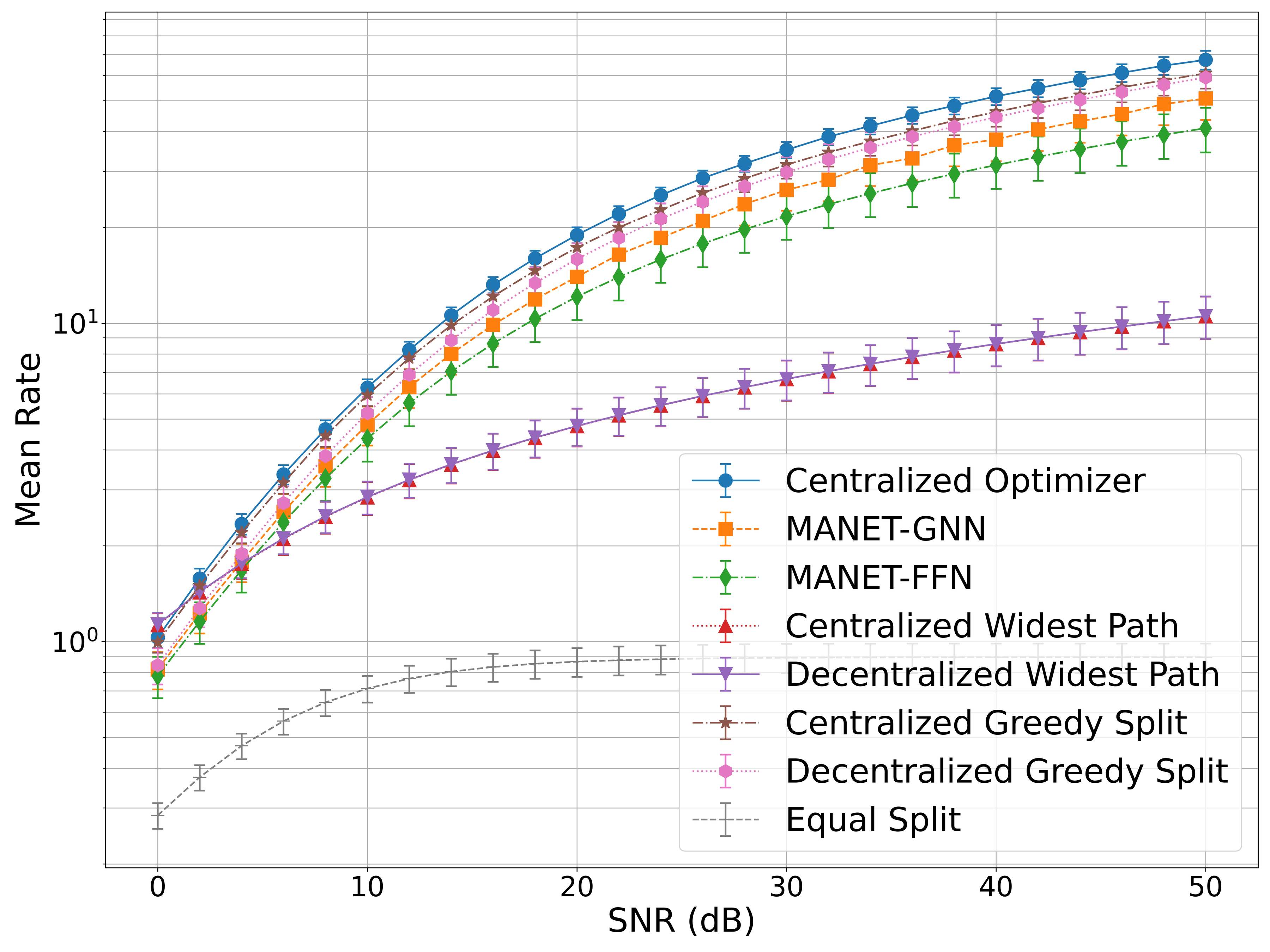}
        \caption{Full \ac{csi}.}
        \label{fig:full_csi_quadriga_unicast}
    \end{subfigure}
    \hfill
    \begin{subfigure}[t]{0.43\textwidth}
        \centering
        \includegraphics[width=\linewidth]{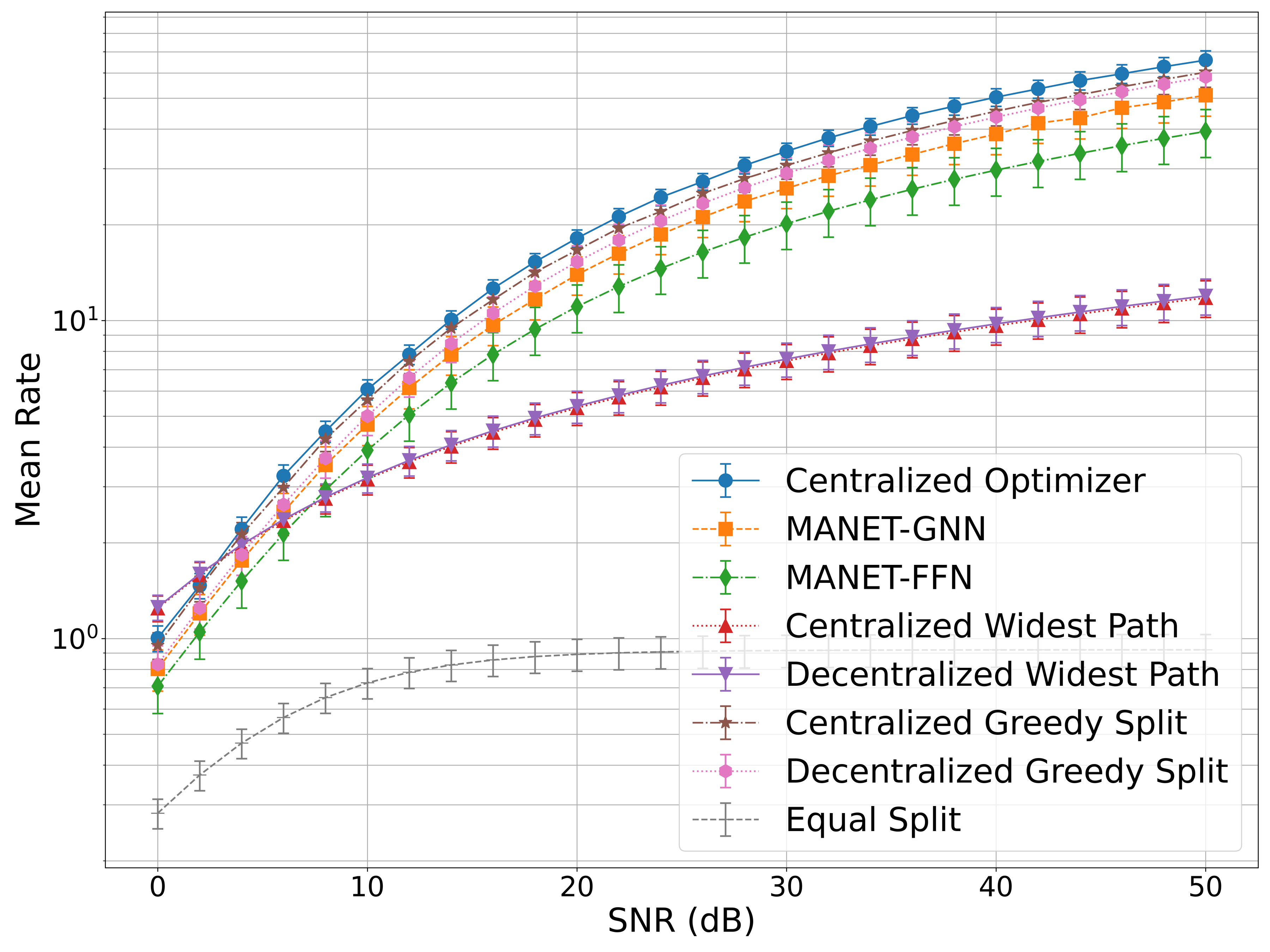}
        \caption{Estimated \ac{csi}.}
        \label{fig:estimated_csi_quadriga_unicast}
    \end{subfigure}
    \caption{Mean rate versus \ac{snr} for unicast (\ref{item:unicast}) communication over
    QuaDRiGa channels.}
    \label{fig:quadriga_unicast}
    \vspace{-0.2cm}
\end{figure*}

\subsubsection{Training Details}
\label{sec:training details}
For each framework, MANET-GNN is trained on 2000 randomly generated topologies, each with $|\mySet{V}|=10$ nodes and $B=6$ frequency bands, of which 1600 are used for training and 400 for validation, with unseen validation samples. Each topology is associated with a fixed channel realization, and the training is performed jointly with \ac{snr} values ranging from 0 to 50 \, dB in increments of 5 \, dB. The model consists of three gated \ac{gnn} layers (total $L=6$ communication rounds), trained for 100 epochs using AdamW (learning rate $5^{-4}$, weight decay $3\times10^{-5}$) with cosine scheduling and dropout ($p=0.2$).

\subsubsection{Testing Details}
\label{sec:testing details}
All reported results are evaluated on 500 independently generated test topologies, each with $|\mathcal{V}|=10$ nodes and $B=6$ frequency bands. For communication frameworks with multiple messages, we use $K=4$. The reported performance is averaged over all test realizations, and the error bars in the figures denote the corresponding 95\% confidence intervals.

\begin{table*}
\centering
\large
\caption{Online complexity of considered algorithms in floating point operations and in communications}
\label{tab:complexity}
\resizebox{\textwidth}{!}{%
\begin{tabular}{lcccl}
\hline
{\centering\textbf{Method} }
& \textbf{Computational Complexity} 
& \textbf{Total Communication} 
& \textbf{Node $i$ Communication} 
& {\centering\textbf{Remarks}} \\
\hline

\textbf{MANET-GNN (Decentralized)} 
& $\mathcal{O}\!\left(L |\myVec{\theta}|\right)$ 
& $\mathcal{O}\!\left(L|\mathcal{E}|BK\right)$ 
& $\mathcal{O}\!\left(L {\rm deg}(i) BK\right)$ 
& $L$ fixed rounds (latency-bounded learned optimizer) \\

\textbf{Centralized Optimizer \ref{itm:BenchCent}} 
& $\mathcal{O}\!\left(I\big(|\mathcal{E}|BK + C(\mathcal{G})\big)\right)$ 
& $-$
& $-$ 
& Iterative optimization with objective evaluation each iteration \\

\textbf{Equal Split \ref{itm:BenchEqual} (Each Node Locally)} 
& $\mathcal{O}\!\left(|\mathcal{V}|\right)$ 
& $-$ 
& $-$ 
& One-shot local allocation (no routing required) \\

\textbf{Greedy-Split \ref{itm:BenchGreedy} (Centralized)} 
& $\mathcal{O}\!\left(C(\mathcal{G}) + |\mathcal{V}|\right)$ 
& $-$ 
& $-$ 
& Shortest-path/smallest-subgraph selection, equal power on chosen route \\

\textbf{Greedy-Split \ref{itm:BenchGreedyDec} (Decentralized)} 
& $\mathcal{O}\!\left(|\mathcal{V}|^2|\mathcal{E}|\right)$ 
& $\mathcal{O}\!\left(|\mathcal{V}||\mathcal{E}|\right)$ 
& $\mathcal{O}\!\left(|\mathcal{V}|\right)$ 
& Distributed Bellman--Ford-style path discovery~\cite{hutson2007distributed} \\

\textbf{Best Single Channel \ref{itm:BenchSingle} (Centralized)} 
& $\mathcal{O}\!\left(C(\mathcal{G}) + |\mathcal{V}|\right)$ 
& $-$ 
& $-$ 
& Max--min link evaluation per band, full power on the strongest bottleneck \\

\textbf{Best Single Channel \ref{itm:BenchSingleDec} (Decentralized)} 
& $\mathcal{O}\!\left(|\mathcal{V}|^2|\mathcal{E}|\right)$ 
& $\mathcal{O}\!\left(|\mathcal{V}||\mathcal{E}|\right)$ 
& $\mathcal{O}\!\left(|\mathcal{V}|\right)$ 
& Distributed Bellman--Ford-style path discovery~\cite{hutson2007distributed} \\

\textbf{Feed-Forward Network \ref{itm:ffn} (Centralized)} 
& $\mathcal{O}\!\left(2B|\mySet{V}|^2H+(n_L-2)H^2+BK|\mySet{V}|^2H\right)$ 
& $-$ 
& $-$ 
& Feedforward fully-connected network with $n_L$ linear layers of dimension $H$ \\

\hline
\end{tabular}}
\end{table*}

\vspace{-0.1cm}


\begin{figure*}[!tbp]
    \centering
    \begin{subfigure}[t]{0.43\textwidth}
        \centering
        \includegraphics[width=\linewidth]{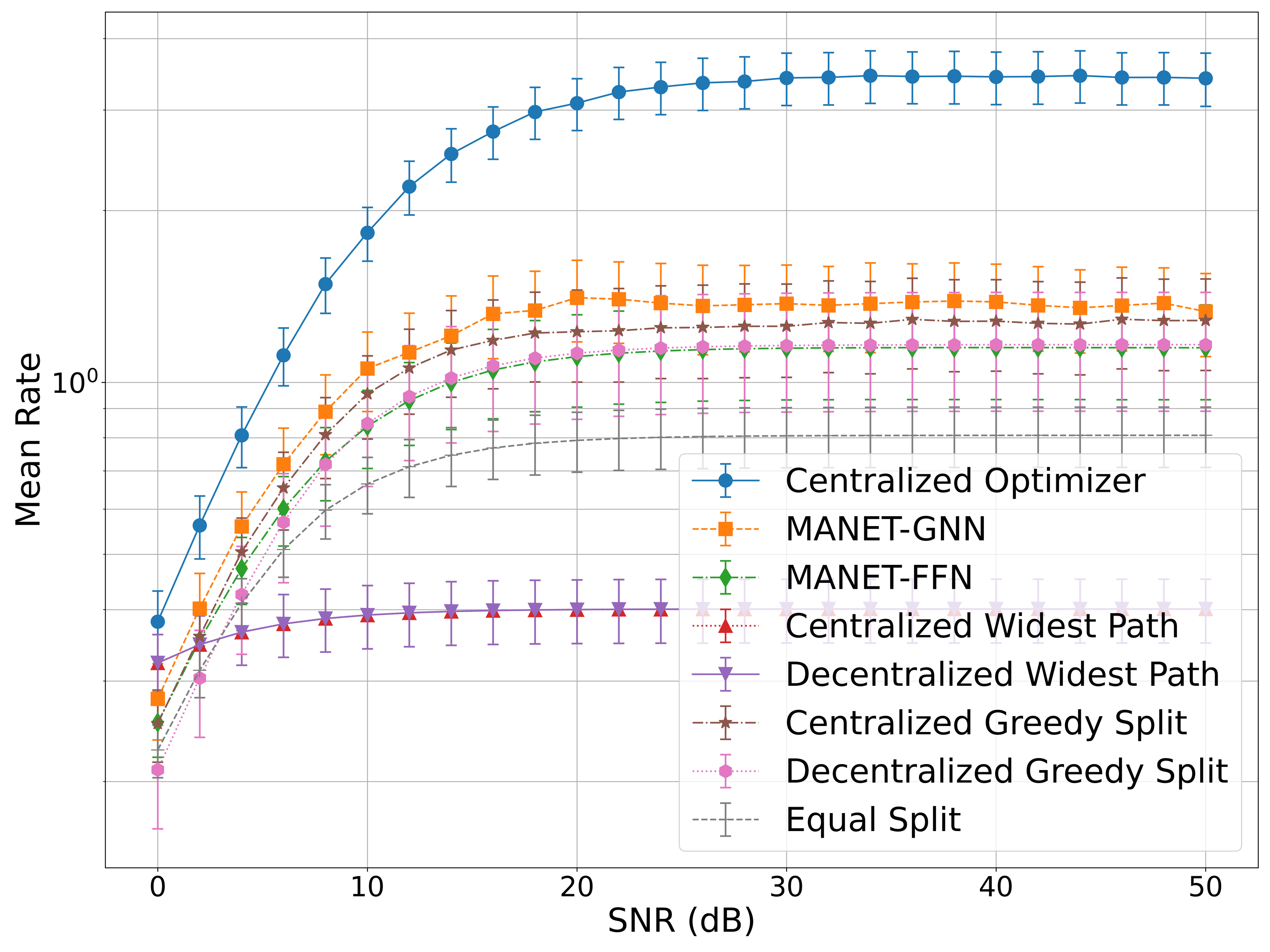}
        \caption{Full \ac{csi}.}
        \label{fig:full_csi_quadriga_multicast}
    \end{subfigure}
    \hfill
    \begin{subfigure}[t]{0.43\textwidth}
        \centering
        \includegraphics[width=\linewidth]{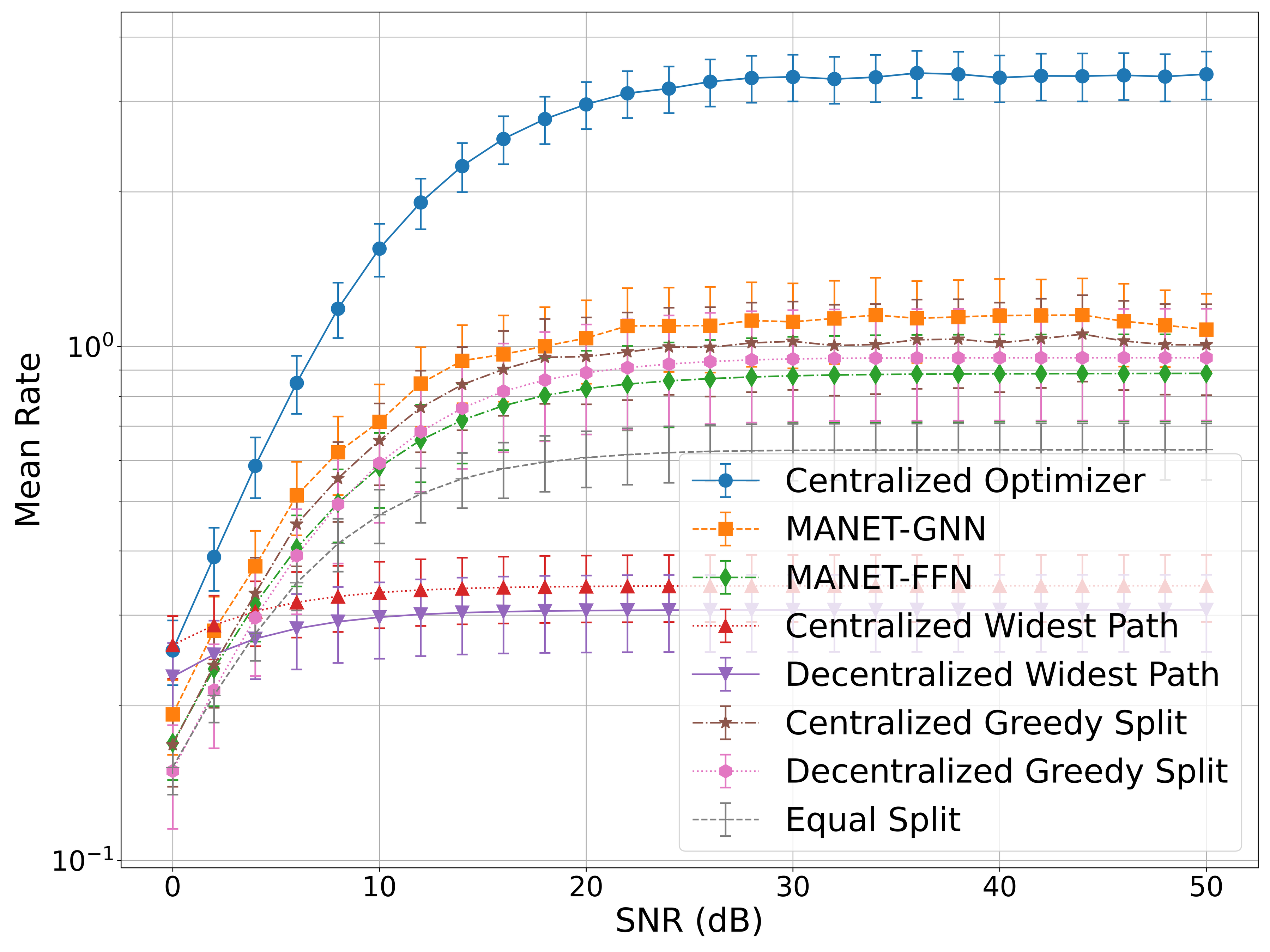}
        \caption{Estimated \ac{csi}.}
        \label{fig:estimated_csi_quadriga_multicast}
    \end{subfigure}
    \caption{Mean rate versus \ac{snr} for multicast communication
    over QuaDRiGa channels.}
    \label{fig:quadriga_multicast}
    \vspace{-0.2cm}
\end{figure*}

\begin{figure*}[!tbp]
    \centering
    \begin{subfigure}[t]{0.43\textwidth}
        \centering
        \includegraphics[width=\linewidth]{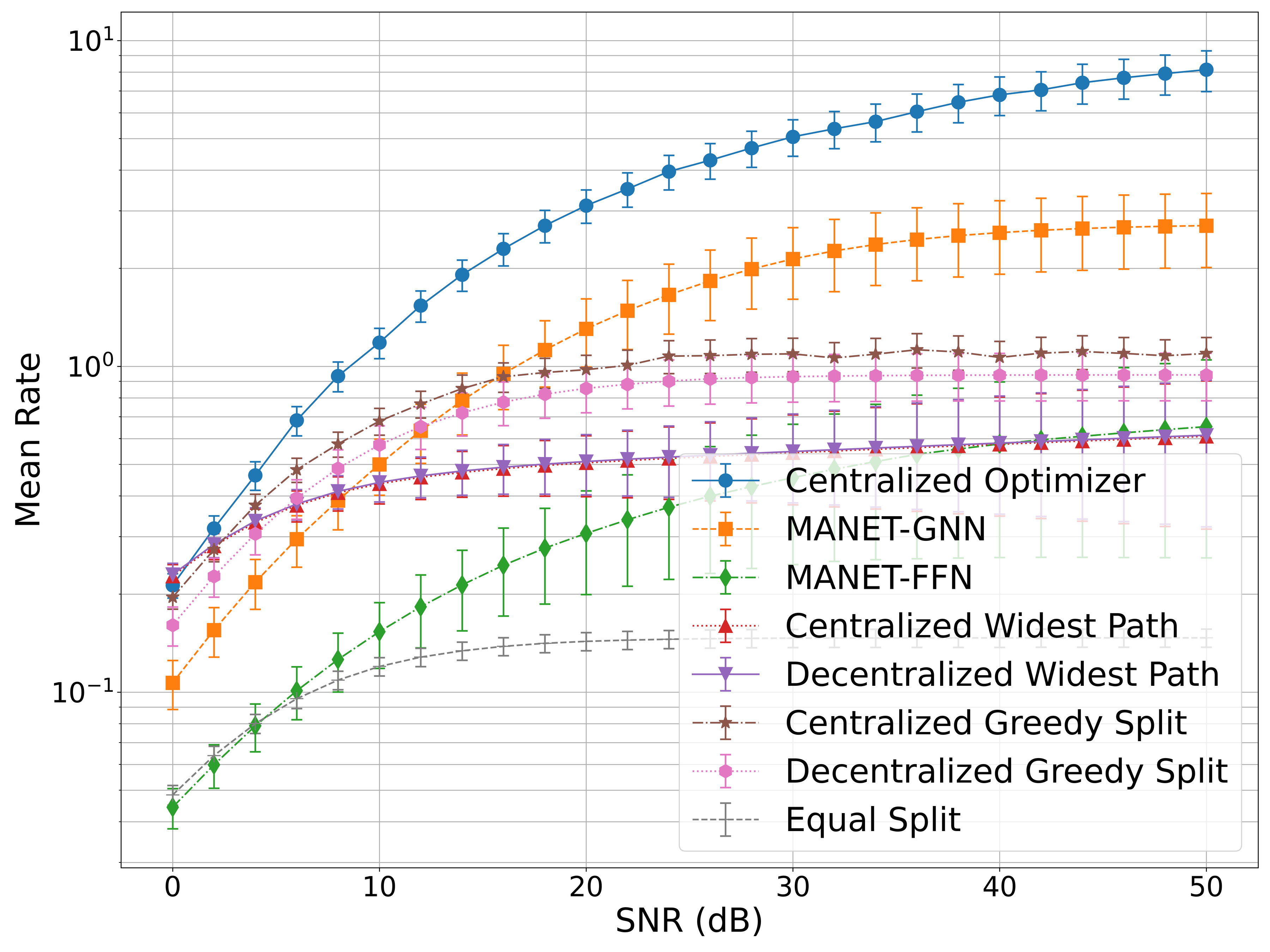}
        \caption{Full \ac{csi}.}
        \label{fig:full_csi_quadriga_multicommodity}
    \end{subfigure}
    \hfill
    \begin{subfigure}[t]{0.43\textwidth}
        \centering
        \includegraphics[width=\linewidth]{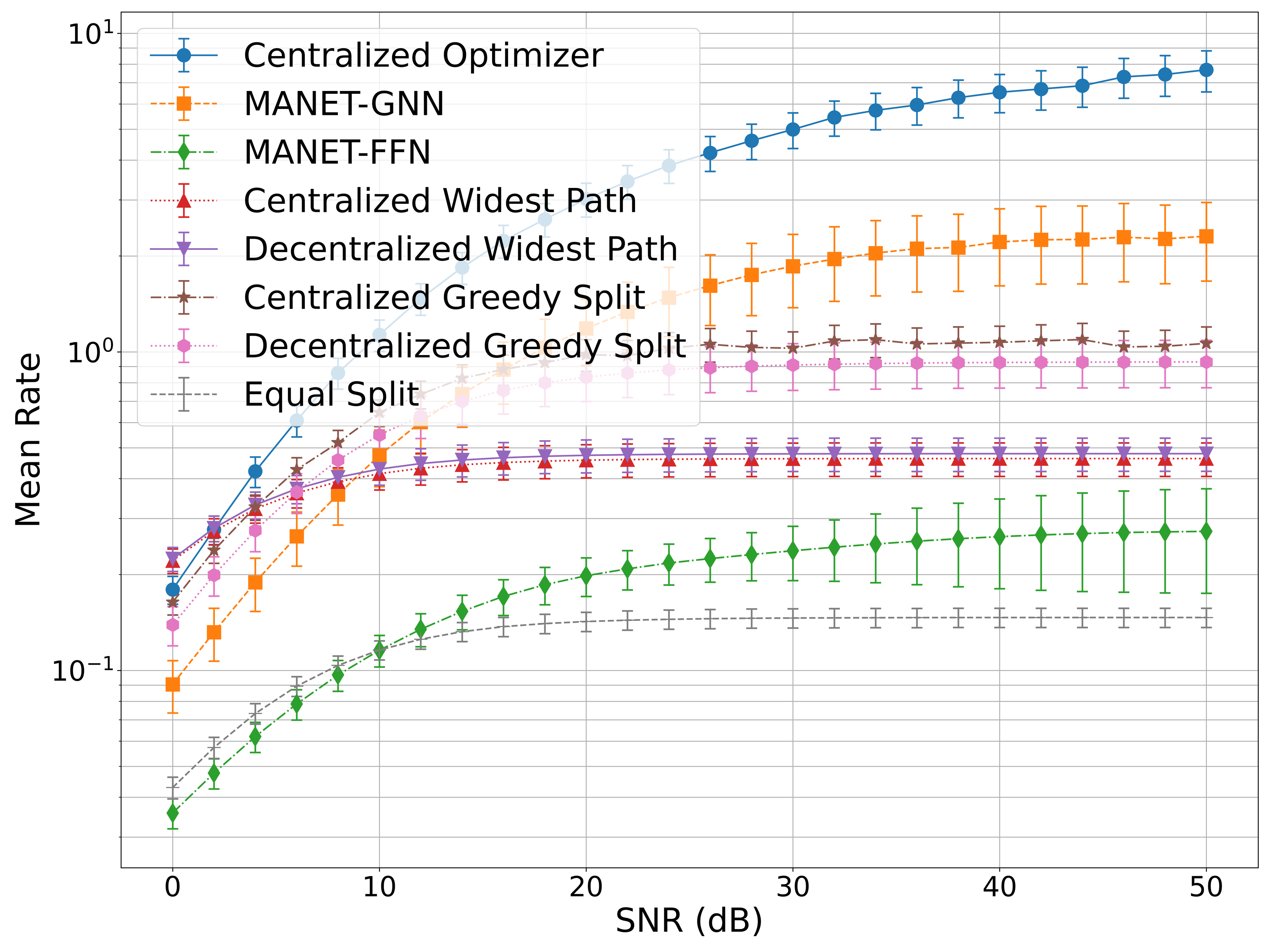}
        \caption{Estimated \ac{csi}.}
        \label{fig:estimated_csi_quadriga_multicommodity}
    \end{subfigure}
    \caption{Mean rate versus \ac{snr} for multicommodity (\ref{item:multicom}) communication
    over QuaDRiGa channels.}
    \label{fig:quadriga_multicommodity}
    \vspace{-0.2cm}
\end{figure*}

\begin{figure*}[!tbp]
    \centering
    \begin{subfigure}[t]{0.43\textwidth}
        \centering
        \includegraphics[width=\linewidth]{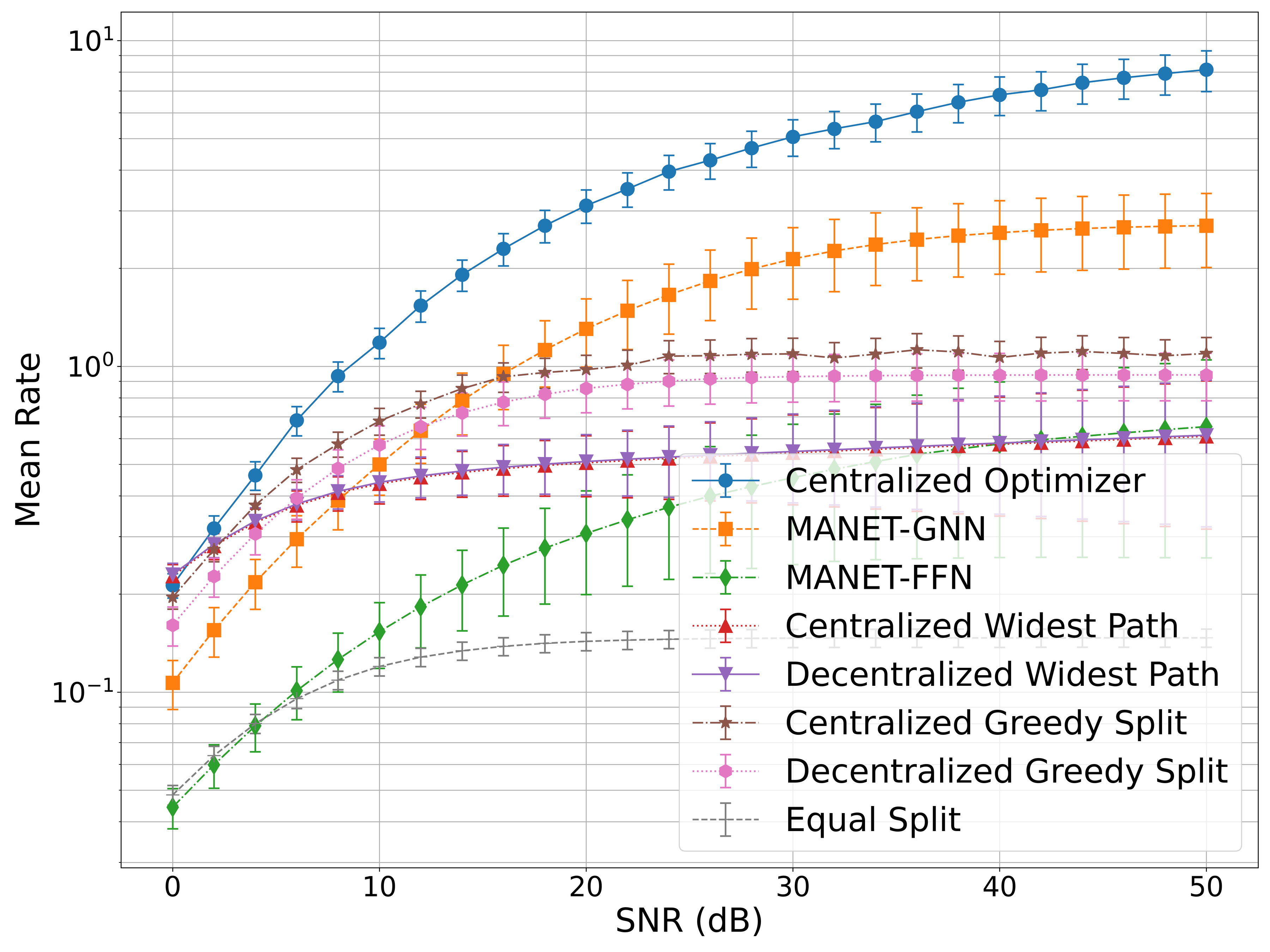}
        \caption{Full \ac{csi}.}
        \label{fig:full_csi_quadriga_convergecast}
    \end{subfigure}
    \hfill
    \begin{subfigure}[t]{0.43\textwidth}
        \centering
        \includegraphics[width=\linewidth]{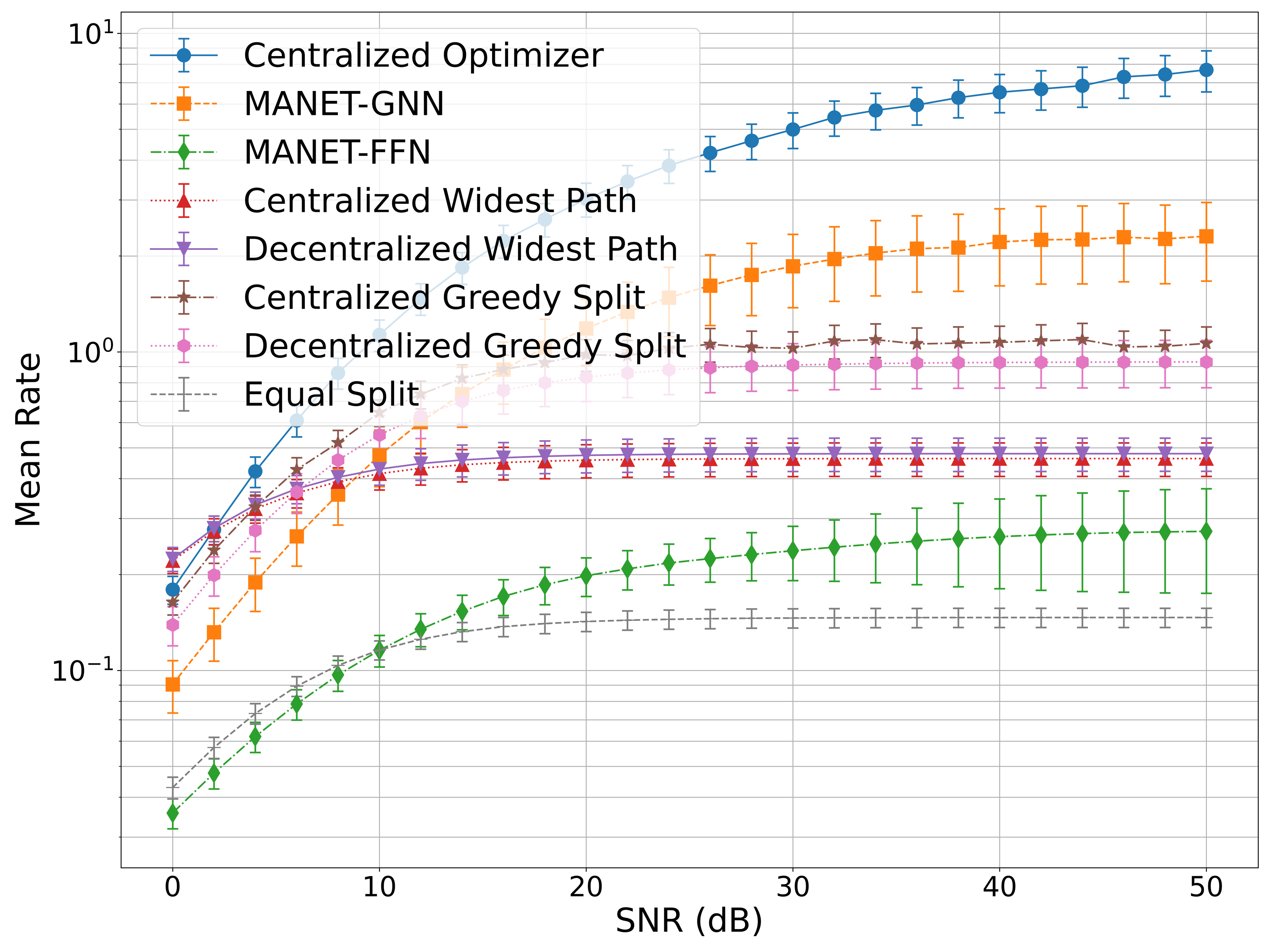}
        \caption{Estimated \ac{csi}.}
        \label{fig:estimated_csi_quadriga_convergecast}
    \end{subfigure}
    \caption{Mean rate versus \ac{snr} for convergecast (\ref{item:convergecast}) communication
    over QuaDRiGa channels.}
    \label{fig:quadriga_convergecast}
    \vspace{-0.2cm}
\end{figure*}

\begin{figure*}[!tbp]
    \centering
    \begin{subfigure}[t]{0.43\textwidth}
        \centering
        \includegraphics[width=\linewidth]{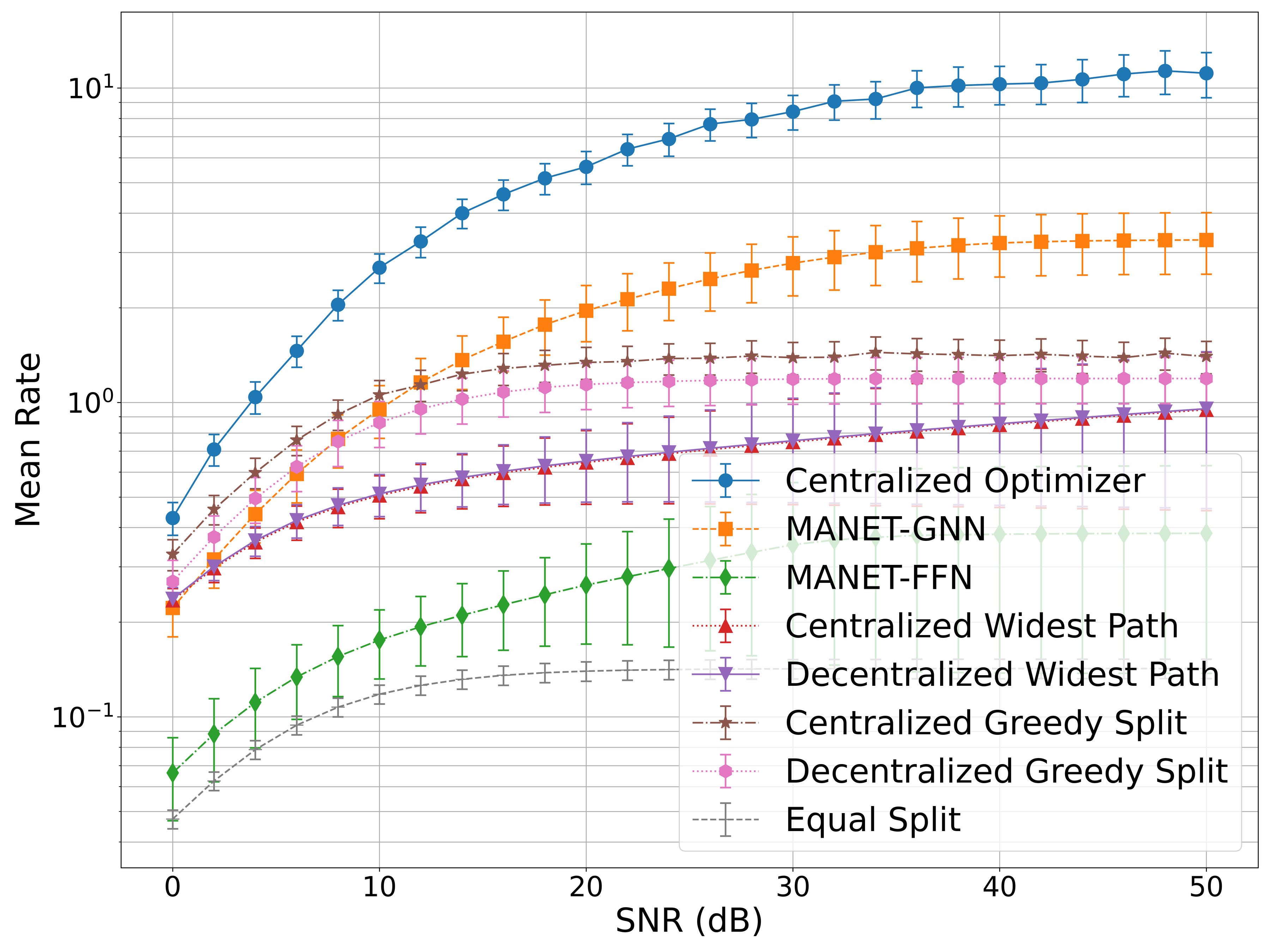}
        \caption{Full \ac{csi}.}
        \label{fig:full_csi_quadriga_many_to_many}
    \end{subfigure}
    \hfill
    \begin{subfigure}[t]{0.43\textwidth}
        \centering
        \includegraphics[width=\linewidth]{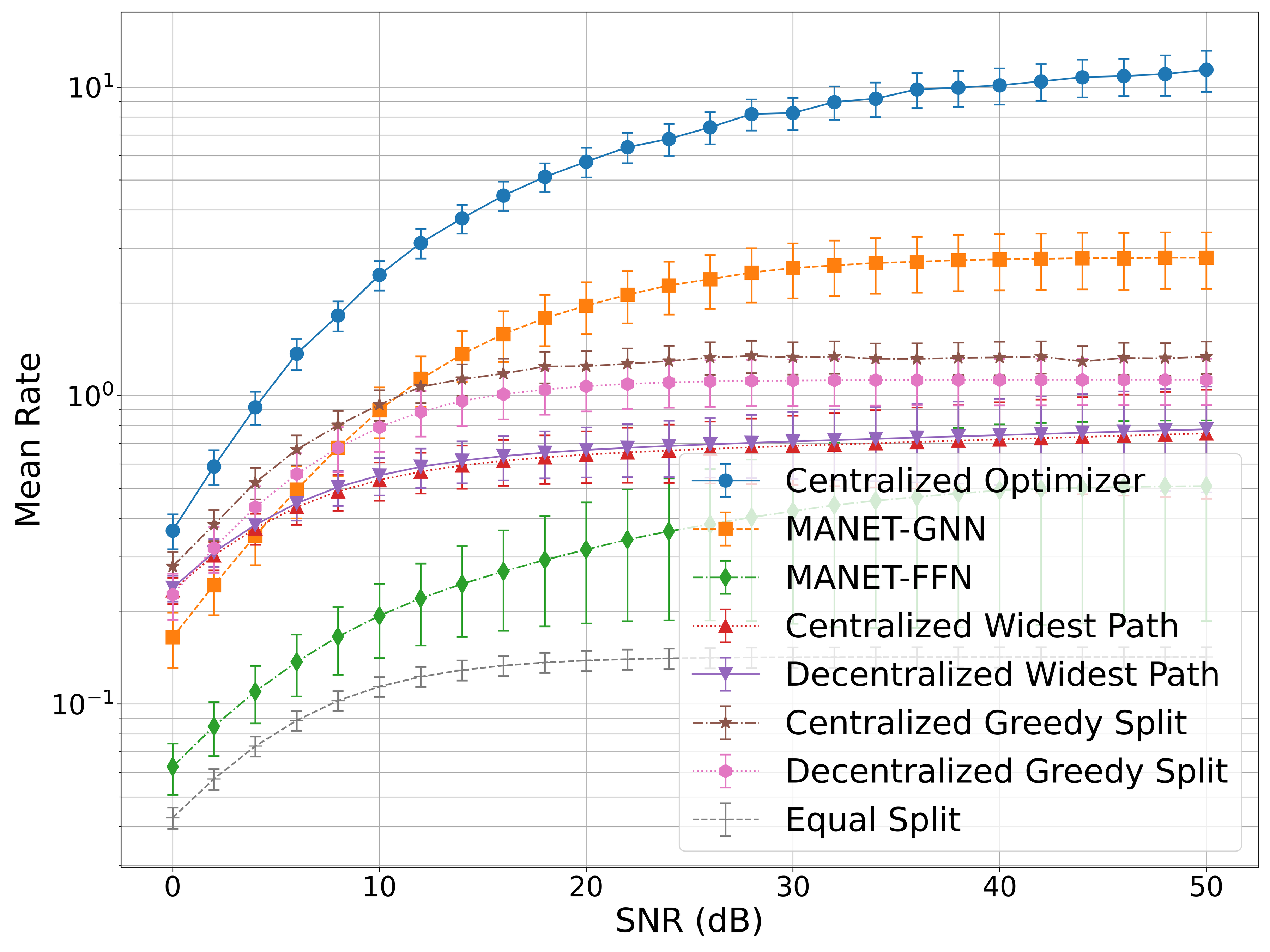}
        \caption{Estimated \ac{csi}.}
        \label{fig:estimated_csi_quadriga_many_to_many}
    \end{subfigure}
    \caption{Mean rate versus \ac{snr} for Many-to-Many (\ref{item:many2many}) communication
    over QuaDRiGa channels.}
    \label{fig:quadriga_many_to_many}
    \vspace{-0.2cm}
\end{figure*}

\subsection{Compared Algorithms}
\label{sec:benchmarks}
\subsubsection{Benchmarks}

We compare MANET-GNN against a diverse set of centralized, decentralized, heuristic, and learned baselines.
\begin{enumerate}[label={B\arabic*}]
    \item \label{itm:BenchCent}
    {\em Centralized Optimizer}, which has access to the complete topology and \ac{csi}, and solves \eqref{eqn: best power allocation_fw} using AdamW~\cite{zhuang2022understanding}. We use four initializations, namely random, Greedy-Split, Widest-Path, and MANET-GNN, and run 1000 iterations with line search over $5$ step sizes. After each update, power allocation is projected onto the per-node  constraints.

    \item \label{itm:BenchEqual}
    {\em Equal-Split}, where each transmitter distributes its power uniformly over all feasible outgoing links and frequency bands. This channel-agnostic rule serves as a simple low-complexity baseline~\cite{Jang2003TransmitPowerAdaptation}.

    \item \label{itm:BenchGreedy}
    {\em \ac{cgs}}, which centrally enumerates all feasible paths, or multicast subgraphs, and selects one of the shortest candidates uniformly at random. The available power is then split equally over the active links and frequency bands~\cite{Draves2004MRLQSR, Saad2012TDMAFDMA}.

    \item \label{itm:BenchGreedyDec}
    {\em \ac{dgr}}, the decentralized counterpart of \ac{cgs}. Nodes use distributed Bellman--Ford updates~\cite{hutson2007distributed} to compute minimum-hop routes using only neighbor exchanges. The resulting allocation is evaluated using the same interference-aware objective.

    \item \label{itm:BenchSingle}
    {\em \ac{cwp}}, which centrally examines the feasible paths, or multicast subgraphs, on each frequency band and selects the route whose weakest link has the largest channel gain. Transmission is then performed over the band with the largest bottleneck value~\cite{low2003duality}.

    \item \label{itm:BenchSingleDec}
    {\em \ac{dwp}}, the decentralized counterpart of \ac{cwp}. Nodes execute Bellman--Ford-like max--min updates~\cite{hutson2007distributed} to find bottleneck-maximizing paths using only local information. For multicast and multi-flow settings, this procedure is applied per destination or transmitter--receiver pair, and the resulting paths are combined into the corresponding routing structure.

    \item \label{itm:ffn}
    {\em \ac{ffn}}, which receives the full channel  and predicts the power allocation using a fully connected neural network. Unlike MANET-GNN, it does not explicitly exploit graph topology, and therefore serves as a centralized learned baseline.
\end{enumerate}
All benchmarks output a power allocation tensor compatible with the parent problem formulation and are evaluated using the same objective functions and constraints as the proposed method.  

\subsubsection{Complexity Analysis}
Table~\ref{tab:complexity} summarizes the online computational and
communication complexities of the considered methods. Let
$C(\mathcal G)$ denote the number of routing objects examined for a given
instance: simple source--destination paths for path-based frameworks and
connected source--receiver subgraphs for multicast. In the worst case,
these quantities scale as
$\mathcal{O}((|\mathcal V|-2)!)$ and
$\mathcal{O}(2^{|\mathcal E|})$, respectively. Consequently, centralized
optimizers and route-enumeration heuristics may become prohibitively
expensive as the network grows or densifies.

MANET-GNN avoids online combinatorial search and performs $L/2$ fixed
message-passing rounds. Its inference complexity is
$\mathcal{O}(\frac{1}{2}L|\myVec{\theta}|)$, with total communication
$\mathcal{O}(L|\mathcal E|BK)$ and per-node communication
$\mathcal{O}(L\deg(i)BK)$. Hence, its latency and signaling overhead are
fixed independently of the number of feasible paths or subgraphs.
Path/subgraph enumeration is required only during offline training and
does not affect deployment complexity.

\vspace{-0.1cm}
\subsection{Comparative Study}
\label{sec:comparative}

We evaluate   MANET-GNN against  \ref{itm:BenchCent}-\ref{itm:BenchSingle} under the different communication frameworks~\ref{item:unicast}-\ref{item:many2many}.
For each framework, we report the average end-to-end rate versus \ac{snr}, averaged over independently generated \ac{manet} realizations.
All evaluations were performed on a $|\mathcal{V}|=10$ nodes and $B=6$ frequency bands \ac{manet}. For multicast and multicommodity, the \ac{manet} includes 4 receivers; in convergecast , there are 4 transmitters; and in  many-to-many, there are 4 transmitter-receiver pairs.

\subsubsection{Unicast}
\label{sec:unicast_results}

We begin with unicast communication (\ref{item:unicast}) over frequency-selective QuaDRiGa channels. Fig.~\ref{fig:quadriga_unicast} shows the results under full and estimated \ac{csi}. The centralized optimizer achieves the highest rate, while the Greedy-Split benchmarks slightly outperform MANET-GNN. This is expected, since unicast involves a single source--destination pair, for which selecting a short route and splitting power along it is often near-optimal. Still, MANET-GNN remains close to these specialized heuristics and preserves its performance under \ac{csi} errors. The weaker performance of Widest-Path, Equal-Split, and MANET-\ac{ffn} reflects their limited ability to exploit topology and band resources.

\subsubsection{Multicast}
\label{sec:multicast_results}

We next consider multicast communication (\ref{item:multicast}), where a common message is delivered from one source to multiple receivers. As shown in Fig.~\ref{fig:quadriga_multicast}, MANET-GNN achieves the best performance among the all benchmarks under both full and estimated \ac{csi}, second only to the centralized optimizer.. Unlike unicast, multicast requires selecting a connected subgraph that reaches all destinations while coordinating power over shared and destination-specific links. The Greedy-Split policies are competitive but do not jointly account for channel quality, interference, and power coupling across the multicast subgraph, causing them to saturate below MANET-GNN at moderate and high \ac{snr}. The similar ordering under estimated \ac{csi} demonstrates the robustness of the learned decentralized policy.

\subsubsection{Multicommodity}
\label{sec:multicommodity_results}

We next consider multicommodity communication (\ref{item:multicom}), where a common source transmits distinct messages to multiple destinations under a shared power budget. Fig.~\ref{fig:quadriga_multicommodity} shows that the centralized optimizer achieves the highest rate, while MANET-GNN is the strongest remaining method from moderate \ac{snr} onward. In this setting, independently selecting short or high-bottleneck routes cannot coordinate the messages competing for the source power and shared network resources. Consequently, Greedy-Split and Widest-Path saturate earlier, whereas MANET-GNN continues to benefit from increasing \ac{snr}. The gap to the centralized optimizer reflects its access to global information, multiple initializations, and iterative line-search optimization, compared with the fixed number of local message-passing rounds used by MANET-GNN.

\subsubsection{Convergecast}
\label{sec:convergecast_results}

We next consider convergecast communication (\ref{item:convergecast}), where multiple sources transmit distinct messages to a common destination. This setting mirrors multicommodity communication, with routes converging toward a shared sink instead of diverging from a shared source. Accordingly, Fig.~\ref{fig:quadriga_convergecast} exhibits similar trends: the centralized optimizer performs best, while MANET-GNN clearly outperforms the remaining benchmarks from moderate \ac{snr} onward. The heuristic policies cannot effectively coordinate commodities that compete for shared links and frequency resources near the destination. In contrast, MANET-GNN captures this coupling through local message passing and remains robust under estimated \ac{csi}.

\subsubsection{Many-to-Many}
\label{sec:many_to_many_results}

Finally, we consider many-to-many communication (\ref{item:many2many}), where multiple source--destination pairs simultaneously transmit distinct messages. This is the most general setting considered, as commodities have different sources and destinations while competing for overlapping routes, frequency resources, and transmit power. As shown in Fig.~\ref{fig:quadriga_many_to_many}, MANET-GNN becomes the strongest benchmark after the low-\ac{snr} regime under both full and estimated \ac{csi}, second only to the centralized optimizer. Unlike unicast, independently selecting short or high-bottleneck routes cannot coordinate the interactions among several simultaneous flows, causing Greedy-Split and Widest-Path to saturate earlier. The consistent behavior under estimated \ac{csi} shows that MANET-GNN preserves its coordination capability despite channel-estimation errors.

\subsubsection{Generalization Across Topologies and Message-Passing Depth}
\label{sec:generalization_results}

Finally, we examine MANET-GNN under unseen and time-varying topologies, and study the effect of the message-passing depth. We first evaluate size generalization in the \ref{item:many2many} framework by comparing a model trained on $|\mathcal V|=10$ graphs, whose average diameter is $3.4$, with $L/2=3$ gated layers, against a model trained on $|\mathcal V|=30$ graphs, whose average diameter is $5.2$, with $L/2=5$ gated layers. Both models are evaluated on the same $200$ unseen $30$-node test graphs.

\begin{figure*}[!t]
    \centering
    \begin{subfigure}{0.43\textwidth}
        \centering
        \includegraphics[width=\textwidth]{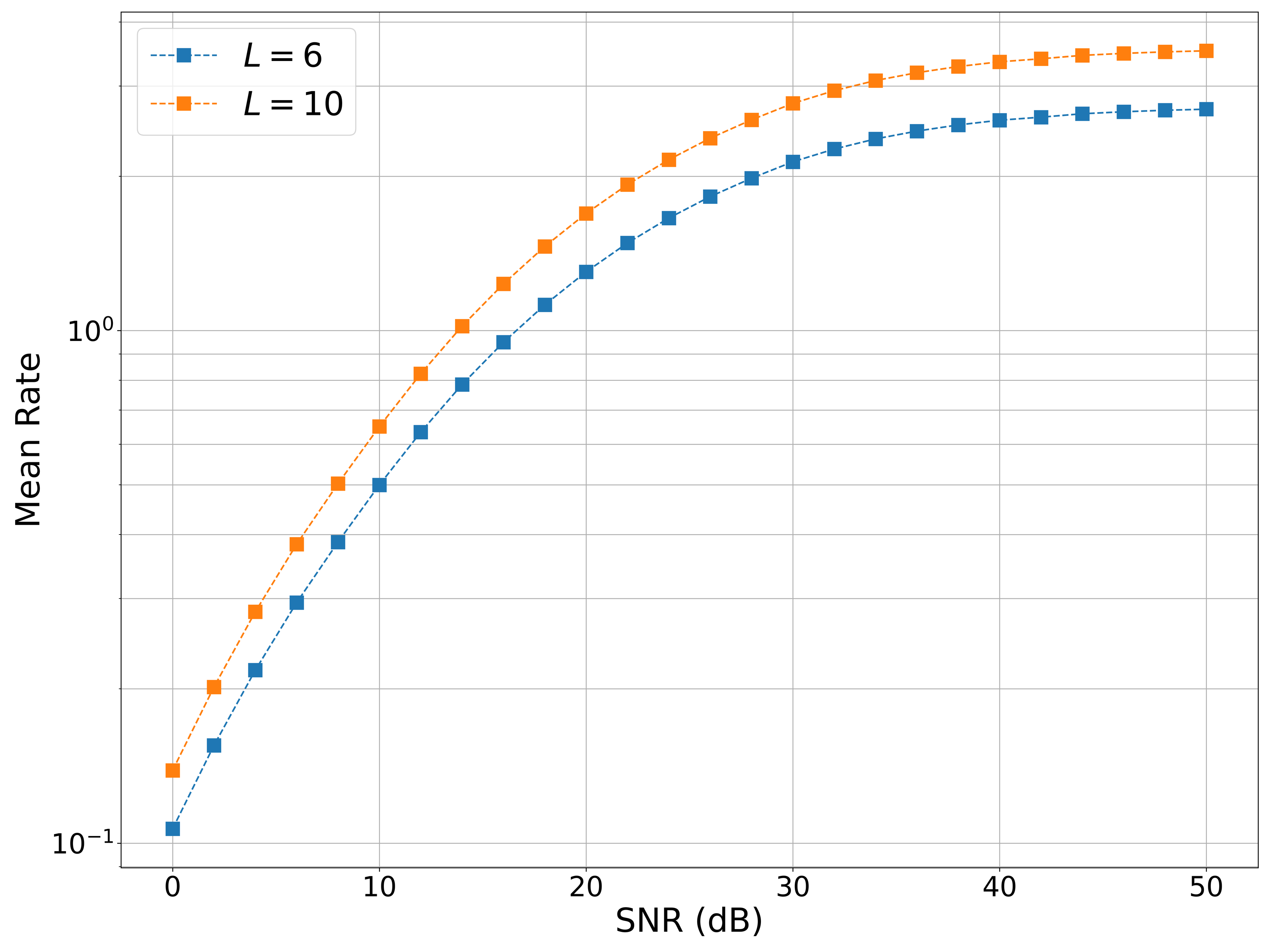}
        \caption{Size generalization.}
        \label{fig:network_size_generalization}
    \end{subfigure}
    \hfill
    \begin{subfigure}{0.43\textwidth}
        \centering
        \includegraphics[width=\textwidth]{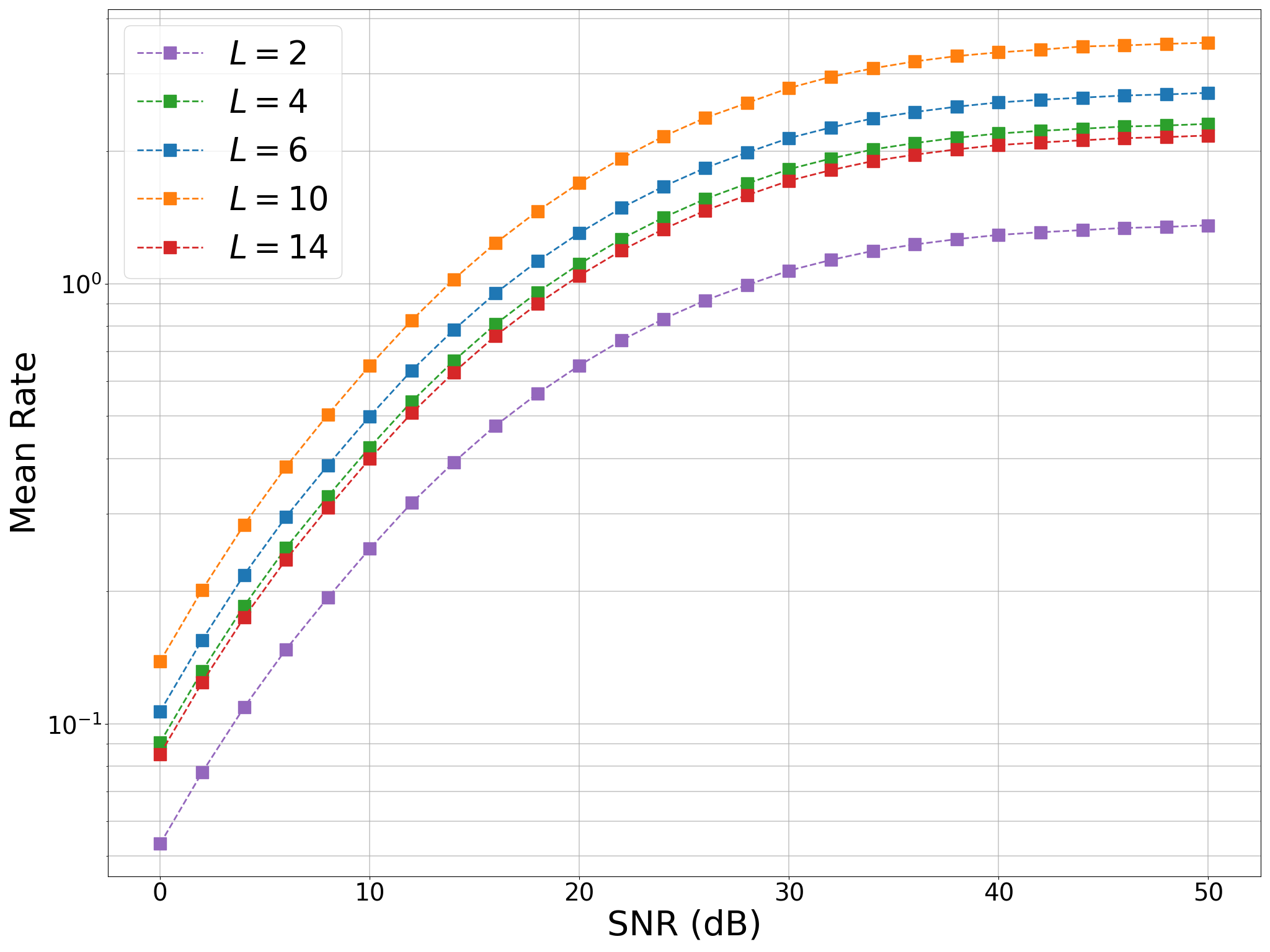}
        \caption{Depth ablation.}
        \label{fig:L_ablation}
    \end{subfigure}
    \caption{Generalization and depth analysis for MANET-GNN in the many-to-many framework.}
    \label{fig:generalization_and_L_ablation}
    \vspace{-0.2cm}
\end{figure*}

As shown in Fig.~\ref{fig:network_size_generalization}, the model trained only on $10$-node graphs remains effective when applied, without retraining, to networks three times larger. This demonstrates that the shared local computations of MANET-GNN generalize across network sizes and connectivity patterns. The model trained directly on $30$-node graphs with $L=10$ achieves about $20\%$ higher mean rate. This is expected, since its five-hop receptive field better matches the test-graph diameter, whereas the $L/2=3$ model observes only three-hop neighborhoods. Thus, MANET-GNN can generalize to larger unseen networks, while matching the message-passing depth to the expected graph diameter further improves performance.

We next isolate the effect of the depth parameter $L$. For this ablation, all MANET-GNN models are trained on  $|\mathcal V|=30$ topologies with average graph diameter $5.2$, and are evaluated on $200$ unseen $30$-node test graphs. Thus, the only architectural parameter varied across the curves is the number of message-passing rounds.
Fig.~\ref{fig:L_ablation} shows that increasing $L$ initially improves performance, since a larger receptive field enables coordination over longer routes and better captures inter-flow coupling. The best performance is obtained for $L=10$, corresponding to $L/2=5$ gated layers, which closely matches the average diameter of the test graphs. Smaller depths, such as $L=2$ and $L=4$, provide only limited local information and therefore perform worse. The $L=6$ model achieves strong performance but remains below $L=10$, reflecting the remaining loss from an insufficient receptive field. Increasing the depth further to $L=14$ degrades the performance, which is consistent with the over-smoothing effect commonly observed in deep \acp{gnn}~\cite{rusch2023survey}. These results demonstrate the expected performance--latency tradeoff: larger $L$ improves coordination up to the point where the receptive field matches the relevant network scale.


\vspace{-0.1cm}
\section{Conclusions}
\label{sec:Conclusions}
We introduced MANET-GNN, a learned decentralized optimization framework for power allocation in multi-channel MANETs. 
By casting the resource allocation task as a unified optimization problem encompassing multiple communication paradigms, MANET-GNN learns topology-aware message-passing policies that operate using only local CSI and a limited number of neighbor exchanges. 
We numerically show that MANET-GNN consistently approaches  centralized optimization across diverse settings while being robust to noisy \ac{csi}. 

\vspace{-0.1cm}

\bibliographystyle{IEEEtran}
\bibliography{IEEEabrv,refs}

\end{document}